\documentclass[a4paper,fleqn]{cas-dc}

\usepackage[authoryear]{natbib}
\usepackage[nolist]{acronym}
\usepackage{makecell}
\usepackage{amsmath}

\def\tsc#1{\csdef{#1}{\textsc{\lowercase{#1}}\xspace}}
\tsc{WGM}
\tsc{QE}

\begin{document}
\let\WriteBookmarks\relax
\def\floatpagepagefraction{1}
\def\textpagefraction{.001}

\begin{acronym}
    \acro{CAM}{Cambridge University Hospitals}
    \acro{MHA}{Mitera Hospital}
    \acro{RUMC}{Radboud University Medical Center}
    \acro{UKA}{University Hospital Aachen}
    \acro{UMCU}{University Medical Center Utrecht}
    \acro{RSH}{Ribera Hospital}
    \acro{DL}{Deep Learning}
    \acro{AI}{Artificial Intelligence}
    \acro{ML}{Machine Learning}
    \acro{DNN}{Deep Neural Network}
    \acro{CNN}{Convolutional Neural Network}
    \acro{MRI}{Magnetic Resonance Imaging}
    \acro{MR}{Magnetic Resonance}
    \acro{AD}{Anomaly Detection}
    \acro{OOD}{Out-of-Distribution}
    \acro{OD}{Outlier Detection}
    \acro{ND}{Noverlty Detection}
    \acro{OSR}{Open Set Recognition}
    \acro{DCE}{Dynamic Contrast Enhanced}
    \acro{ODELIA}{ODELIA Breast MRI Challenge}
    \acro{DUKE}{Duke Breast Cancer MRI}
    \acro{TCGA-KIRP}{Cancer Genome Atlas Cervical Kidney Renal Papillary Cell Carcinoma Collection}
    \acro{TCGA-LIHC}{Cancer Genome Atlas Liver Hepatocellular Carcinoma Collection}
    \acro{PFMRIP}{Prostate Fused MRI Pathology}
    \acro{ID}{In-Distribution}
    \acro{OOD}{Out-of-Distribution}
    \acro{GAN}{Generative Adversarial Network}
    \acro{VAE}{Variational Autoencoder}
    \acro{VQ-GAN}{Vector Quantized-Generative Adversarial Network}
    \acro{CT}{Computed Tomography}
    \acro{OCT}{Optical Coherence Tomography}
    \acro{HU}{Hounsfield Scale}
    \acro{RA}{Reversed Autoencoder}
    \acro{ROC}{Receiver Operating Characteristic}
    \acro{AUROC}{Area Under the Receiver Operating Characteristic Curve}
    \acro{PE}{Positional Encoding}
    \acro{RA}{Reversed Autoencoder}
    \acro{MSE}{Mean-Squared-Error}
    \acro{KL}{Kullback-Leibler}
    \acro{SI-VAE}{Soft-Introspective-Variational Autoencoder}
    \acro{QA}{Quality Assurance}
    \acro{HDI}{Human Development Index}
    \acro{EUSOBI}{European Society of Breast Imaging}
    \acro{PC}{PatchCore}
    \acro{DDPM}{Denoising Diffusion Probabilistic Model}
\end{acronym}

\shorttitle{Unsupervised Anomaly Detection for Image Dataset Quality Assurance in Multi-Center Breast MRI}

\shortauthors{C. Tappermann et~al.}  

\title [mode = title]{Unsupervised Anomaly Detection for Image Dataset Quality Assurance in Multi-Center Breast MRI}

\author[1]{Chiara Tappermann}[style=western, orcid=0009-0003-4450-098X, bioid=1]
\cormark[1]

\credit{Conceptualization, Investigation, Methodology, Software, Validation, Visualization, Writing – original draft}

\author[1]{Steffen Renisch}[style=western, orcid=0009-0003-7011-2630]
\credit{Conceptualization, Methodology, Project administration, Validation, Writing – review and editing}

\author[1]{Lars Ole Schwen}[style=western,orcid=0000-0003-0195-9603]
\credit{Conceptualization, Methodology, Validation, Writing – review and editing}

\author[1]{Hans Meine}[style=western,orcid=0000-0002-7557-5007]
\credit{Conceptualization, Writing – review and editing}

\author[1, 2]{Horst K. Hahn}[style=western,orcid=0000-0001-7512-5762]
\credit{Validation, Writing – review and editing}

\author[1, 3]{Eike Petersen}[style=western,orcid=0000-0003-0097-3868]
\credit{Conceptualization, Methodology, Supervision, Validation, Writing – review and editing}

\affiliation[1]{organization={Fraunhofer Institute for Digital Medicine MEVIS},
            addressline={Max-Von-Laue-Straße 2}, 
            city={Bremen},
            postcode={28359}, 
            state={Bremen},
            country={Germany}}

\affiliation[2]{organization={University of Bremen},
            addressline={Bibliothekstraße 1}, 
            city={Bremen},
            postcode={28359}, 
            state={Bremen},
            country={Germany}}

\affiliation[3]{organization={Institute of Diagnostic and Interventional Radiology, Hannover Medical School},
            addressline={Carl-Neuberg-Str. 1}, 
            city={Hannover},
            postcode={28359}, 
            state={Lower Saxony},
            country={Germany}}

\cortext[1]{Corresponding author}

\begin{abstract}
Corrupted, inconsistent, or anomalous data silently threatens the safety and reliability of high-risk medical AI. 
Despite growing regulatory recognition of dataset quality assurance (QA) in this field, scalable automated detection remains underdeveloped.

We systematically employ unsupervised anomaly detection (AD) and out-of-distribution (OOD) detection as an automated dataset QA mechanism for multi-center dynamic contrast-enhanced breast MRI. 
We construct a controlled AD benchmark comprising seventeen realistic QA-relevant anomaly types from six public datasets (protocol violations, processing errors, incorrect anatomical regions) and propose a taxonomy of radiological imaging anomalies based on human visual perception, enabling fine-grained AD failure mode analysis. 
The benchmark includes near-, medium-far-, far-OOD samples, in-distribution and external normal data. 
Four methods are evaluated, including a projection-based method extended with a domain-specific feature extractor and a novel positional encoding, a reconstruction-based approach extended to full 3D volumes with an augmented training objective, and two unmodified hybrid OOD detection methods.

Medium-far- and far-OOD samples are detected reliably, whereas near-OOD samples and external normal data from unseen institutions expose method-specific differences.
The 3D reconstruction-based approach achieves the best balances of detection performance (AUROC: 0.936 $\pm$ 0.006) and cross-institutional transferability. The projection-based method with positional encoding achieves the highest overall detection performance (AUROC: 0.954 $\pm$ 0.002).
Both hybrid methods exhibit critical failure modes, confirming that methods validated for one modality or anatomy may not generalize without domain-specific adaptation.
Implants and mastectomies remain an open challenge across all methods.
Our findings establish a foundation and practical guidance on scalable unsupervised QA in medical AI pipelines.
\end{abstract}


\maketitle

\section{Introduction}\label{sec:introduction}

Corrupted, inconsistent, or anomalous data represent a silent but significant threat to the quality, safety, and reliability of medical \ac{AI} systems, as they can propagate undetected through model training and deployment, leading to model failures that can directly compromise patient safety and clinical outcomes~\citep{Herath2025,Petersen2022,Zimmerer2022a}.
Despite the growing regulatory recognition of the importance of data governance, data integrity, and dataset \ac{QA} as prerequisites for safe medical \ac{AI} systems~\citep{Schwabe2024, IEEE2801-2022}, the methodological approaches for automatically detecting such irregularities at scale remain underdeveloped~\citep{Schwabe2024}.
Regulations such as the EU \ac{AI} Act define quality criteria for high-risk \ac{AI} systems that not only concern model performance but also the underlying datasets~\citep{AIAct}. 
Dataset cleaning and error minimization become essential practices to ensure that training, validation, and test data support robust model development~\citep{AIAct, Schwabe2024}. 
This underscores the need for concrete methodological approaches in addition to conceptual frameworks to implement these requirements efficiently in practice.
Existing approaches for dataset quality control in medical imaging are insufficient for this purpose. 
They are not designed to capture semantic image content, are not scalable, focus on per-image scoring, or remain purely conceptual~\citep{Esteban2017, Galbusera2024, Herath2025, Kastryulin2023, Marziali2026, Rooij2024, Schwabe2024}. 

Unsupervised \ac{AD} and \ac{OOD} detection methods offer a practical and scalable solution to this problem.
By learning the distribution of normal \ac{ID} data, they can identify anomalous, corrupted, or distributionally inconsistent samples, addressing the practical limitations of manual verification on large-scale multi-center datasets~\citep{Cote2024a, Yang2024}.
While \ac{AD} and \ac{OOD} detection in imaging has been extensively studied in both industrial and medical contexts, their application has been almost exclusively limited to diagnostic tasks or defect detection. 
They are evaluated on synthetic anomalies, homogeneous datasets, or controlled settings and do not explicitly address the heterogeneity of real-world, multi-center data as a \ac{QA} challenge.
We argue that they can be employed as \ac{QA} mechanisms within this regulatory context to ensure dataset integrity for model development and deployment.

There is a significant gap in understanding how these methods translate and generalize across the heterogeneity of real-world, multi-center breast \ac{MRI} datasets for \ac{QA}, what failure modes emerge under realistic institutional heterogeneity, and what the practical tradeoffs are between detection performance, transferability, and methodological complexity.
We directly address this gap by proposing a systematic benchmark for these challenges using unsupervised \ac{AD} and \ac{OOD} detection as a dataset \ac{QA} mechanism.
Using \ac{DCE} breast \ac{MR} subtraction images as the main data source acquired across multiple centers, protocols, and scanner setups, we evaluate the detection of a broad range of clinically motivated anomalies, including protocol violations, processing errors, and incorrect anatomical regions.
We provide a controlled \ac{AD}/\ac{OOD} detection benchmark, a systematic taxonomy, and domain-specific adaptations of methods for realistic multi-center heterogeneity.
The adapted code and data splits are publicly available\footnote{\url{https://github.com/FraunhoferMEVIS/BreastMRIAnomalyQA}}.
Our main contributions are as follows:

\begin{enumerate}[a)]
    \item \textbf{A benchmark for anomaly and out-of-distribution detection for dataset quality assurance:} We apply unsupervised \ac{AD} and \ac{OOD} detection for image dataset \ac{QA} in the context of development and deployment for medical \ac{AI}, shifting the focus from pathology discovery to the automated identification of data-level irregularities. For the implementation and evaluation, we construct a controlled \ac{AD}/\ac{OOD} detection benchmark derived from a large multi-center breast cancer imaging study, incorporating samples from external datasets, systematic transformations of the original data, and data to model different modalities and anatomies. The dataset covers a wide range of clinically motivated anomaly types across six public datasets. This provides a realistic benchmark for evaluating \ac{AD} and \ac{OOD} detection methods in the context of multi-center heterogeneity.

    \item \textbf{Quality assurance anomaly taxonomy:} Existing ano\-maly categorizations in medical imaging are not designed to capture dimensions relevant to dataset \ac{QA}. We therefore propose a systematic taxonomy of radiological image data anomalies, categorizing them along four clinically motivated dimensions, including \textit{protocol and modality}, \textit{anatomical and structural alterations}, \textit{orientation and field of view}, and \textit{spatial extent}. Each dimension is graded on a three-point scale, and the aggregate grade provides a conceptual foundation for characterizing different anomaly types for medical imaging use cases.

    \item \textbf{Domain-specific adaptation of anomaly detection methods:} We extend two \ac{AD} methods with domain-specific adaptations to enable their application to volumetric \ac{DCE} breast \ac{MRI} and to address failure modes that prevent reliable performance. For the projection-based method, we introduce a domain-specific feature extractor and a novel positional encoding extension to capture spatial context. For the reconstruction-based method, we extend the original method to enable full volumetric processing and augment the training objective to enable sharp and reliable reconstructions.

    \item \textbf{Systematic evaluation of anomaly detection methods:} We evaluate four unsupervised \ac{AD} and \ac{OOD} detection techniques for multi-center breast \ac{MRI}, including a projection-based method, a reconstruction-based method, and two hybrid approaches. Our evaluation reveals method-specific strengths, failure modes, and generalization behavior under realistic multi-center heterogeneity.
\end{enumerate}

\section{Literature Review}

\subsection{Data Quality Assurance in Medical Imaging}
\label{sec:data-qa}

Existing approaches for data quality assurance in medical imaging are insufficient for medical \ac{AI} development and deployment.
Metadata-based checks and histogram analysis can identify a limited class of low-level errors, such as wrong file formats or implausible intensity ranges, but are insensitive to semantically inconsistent samples, subtle pre-processing errors, or distribution inconsistencies that are only apparent in the image content itself~\citep{Galbusera2024, Herath2025}.

Manual radiologist review, while sensitive to content-level irregularities, is not scalable to thousands of samples, especially in multi-center studies.
Existing radiological image quality assessment frameworks primarily focus on radiological, per-image scoring systems or \ac{MRI}-specific image quality metrics ~\citep{Kastryulin2023, Marziali2026, Rooij2024}. These methods are designed to assess individual scans for acquisition quality and diagnostic usability, not to capture dataset-level consistency, distributional shifts, or pre-processing-related irregularities.

Automated quality control pipelines have been proposed that predict per-scan usability from extracted image quality metrics, most prominently for brain \ac{MRI}~\citep{Esteban2017}.
Such pipelines demonstrate that image-based quality control can be automated at scale, but they are typically supervised, require expert-labeled quality ratings, and are adapted to a single modality and anatomical region. 
To identify other anomalies, additional labeled data would have to be provided, and the models would have to be retrained.

Similarly, the METRIC framework introduced by~\citet{Schwabe2024} provides an important conceptual foundation for data integrity and quality in medical \ac{AI}, proposing a structured set of data quality dimensions for trustworthy medical \ac{AI} in a specific clinical use case, rather than per-image radiological quality. 
Motivated by regulatory and trustworthiness considerations, it nonetheless offers only a conceptual blueprint rather than algorithmic methods for identifying anomalous or inconsistent samples or for evaluating these methods under realistic clinical conditions~\citep{Schwabe2024}.

In summary, none of these approaches are designed to capture the semantic content of the image at scale, are sufficiently scalable, move beyond a purely conceptual framework, or detect the full range of data-level irregularities that can cause harm during \ac{AI} model development and deployment in heterogeneous multi-center settings.

\subsection{Unsupervised Anomaly and Out-of-Distribution Detection}

Unsupervised \ac{AD} and \ac{OOD} detection methods offer a practical and scalable solution to this problem.
Rather than requiring annotated examples of both normal and typically scarce anomalous samples, they learn only the distribution of normal \ac{ID} data and detect anomalous, corrupted, or distributionally inconsistent samples that deviate from it~\citep{Yang2024}. 
Therefore, it is essential to establish a clear and precise definition of what is considered normal or \ac{ID} for a specific task, since it is not possible to identify and model every potential type of anomaly in advance~\citep{Yang2024}.
This concept lets such methods generalize to unseen anomaly types~\citep{Cai2025, Cui2023}.

During data curation, \ac{AD} and \ac{OOD} detection methods enable the identification of anomalous samples, addressing the practical limitations of manual verification on large-scale multi-center datasets~\citep{Cote2024a}. 
During deployment, they provide a safeguard by detecting inputs that deviate from the training distribution and may lead to unreliable or overconfident predictions, compromising patient care~\citep{Zimmerer2022a}.
Although in the medical domain these concepts are commonly applied to rare disease recognition and health screening tasks~\citep{Cai2025}, we argue that they can be employed as \ac{QA} mechanisms within this regulatory context to ensure dataset integrity for model development and deployment.

\subsection{Application in Medical and Industrial Imaging}

Unsupervised \ac{AD} and \ac{OOD} detection methods have been applied across various domains, particularly in industrial manufacturing and increasingly in medical imaging~\citep{Bao2024a, Cai2025, Liu2024, Roth2022a}.
In industrial settings, these methods are primarily applied to manufacturing defect detection~\citep{Roth2022a}.
In the medical domain, a variety of use cases have been explored across a wide range of imaging modalities and anatomical regions, including brain \ac{MRI}, head \ac{CT}, liver \ac{CT}, retinal \ac{OCT}, chest X-ray, histopathology, pelvic \ac{MRI} and colonoscopy predominantly for diagnostic purposes such as rare disease recognition or pathology detection ~\citep{Bao2024a, Bercea2024, Bercea2024a, Cai2025, Graham2022a, Graham2023, Kadhim2026, Tschuchnig2022}.

In breast cancer imaging specifically, \ac{AD} and \ac{OOD} detection methods have been proposed for mammography, ultrasound, and \ac{MRI}~\citep{Lang2023, Oviedo2025, Tschuchnig2022, Zhang2025}.
\citet{Oviedo2025} and~\citet{Lang2023} apply an explainable \ac{AD} model and a reconstruction-based \ac{AD} model, respectively, to \ac{DCE} breast \ac{MRI} for cancer detection. 
Both have a purely diagnostic focus, framing malignant lesions as the anomaly of interest rather than data-level irregularities.

Existing applications remain almost exclusively studied in diagnostic use cases in medicine and defect detection in industry.
These methods are typically developed and evaluated on synthetic anomalies, homogeneous datasets, or controlled settings, and do not explicitly address the heterogeneity of real-world, multi-center data as a \ac{QA} challenge.
In contrast, the use of \ac{AD} and \ac{OOD} detection as an automated \ac{QA} mechanism to identify data-level irregularities is a fundamentally different problem that has received little systematic attention in the literature.

\subsection{Anomaly and Out-of-Distribution Detection Method Categories}
\label{sec:ad-categories}

The literature proposes two broad categories of deep unsupervised \ac{AD} approaches: reconstruction-based and pro\-jection-based methods, the latter also being referred to as feature embedding-based or feature reference-based methods~\citep{Bao2024a, Cai2025, Liu2024}. 
Reconstruction-based methods detect anomalies by measuring deviations between the input and its reconstructed pseudo-normal representation, with larger reconstruction errors indicating a mismatch with the learned normal data distribution~\citep{Bao2024a, Bercea2024, Cai2025}. 
This can include image reconstruction and feature reconstruction~\citep{Cai2025}.
These methods employ architectures such as \acp{GAN}, \acp{VAE}, transformers, and diffusion models and train them on normal data~\citep{Bao2024a, Liu2024}.
In contrast, projection-based methods map data into an embedding space to enhance the separability of normal and anomalous samples. 
Projection-based methods include different one-class classification methods, teacher-student architectures, memory bank-based methods, normalizing flow methods, and distribution map-based approaches~\citep{Bao2024a, Liu2024}.

\subsection{Breast Cancer Screening}

Breast cancer is one of the most common cancers among women worldwide. 
In 2022, breast cancer affected 2.3 million women globally, causing 670,000 deaths. 
Depending on the \ac{HDI}, up to one in twelve women will be diagnosed with breast cancer over the course of their lifetime~\citep{WHO2026Breast}.
Although the primary imaging modality for breast cancer screening is mammography, the \ac{EUSOBI} recommends \ac{MRI} as an additional screening tool for certain conditions or characteristics, e.g., women with dense breast tissue~\citep{Mann2022}.
However, the examination of breast \ac{MRI} is more time-consuming compared to reading mammograms and increases radiologist workload, amplifying the need for \ac{AI}-assisted tools~\citep{MuellerFranzes2025a}.

At the same time, \ac{DCE} breast \ac{MRI} exhibits substantial inter‑center variability originating from different acquisition protocols, hardware manufacturers, contrast administration, and reconstruction settings in addition to variations in patient anatomy.
This technical and biological heterogeneity complicates the definition of consistent normative data distributions, making it a highly relevant but particularly challenging use case for unsupervised \ac{AD} and \ac{OOD} detection methods.
Nevertheless, the appropriate representation of the variation in the training data is crucial for training medical \ac{AI} models that are reliable and robust in practice.

\section{Material and Methods}\label{sec:material_background}

\begin{figure*}
    \centering
    \includegraphics[width=\textwidth]{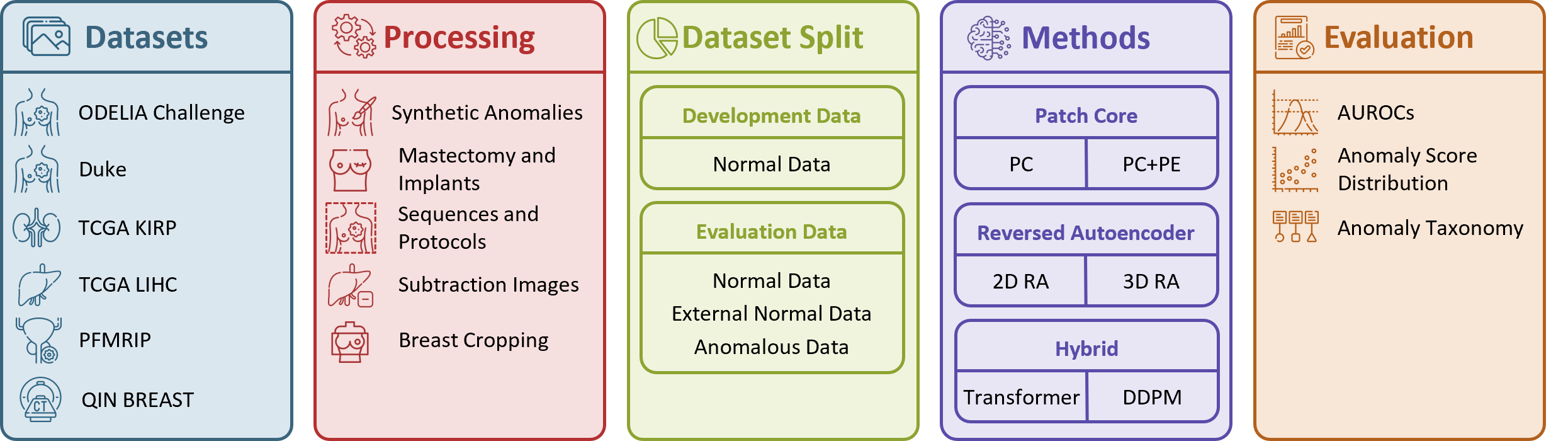}
    \caption{Overview of the experimental pipeline. (1) Data sources, comprising six publicly available medical imaging datasets: the ODELIA Challenge dataset for normal, external, and anomalous data, the Duke dataset for external normal data, and TCGA KIRP, TCGA LIHC, PFMRIP, and QIN BREAST for anomalous data. (2) Data processing steps, including generation of synthetic anomalies, filtering of structural alterations, protocol-specific sequences, subtraction image generation for additional anatomical structures, and breast cropping. (3) Dataset split into development data (normal training and validation data) and evaluation data (normal, external, and anomalous data). (4) Evaluated methods, including PatchCore with and without Positional Encoding (PC, PC+PE), Reversed Autoencoder (2D RA, 3D RA), and two Transformer-based and Denoising Diffusion Probabilistic Model-based hybrid OOD detection approaches. (5) Evaluation including anomaly classification AUROCs, anomaly score distributions, and anomaly taxonomy grading.}
    \label{fig:pipeline}
\end{figure*}

To assess whether unsupervised \ac{AD} and \ac{OOD} detection methods can serve as a scalable dataset \ac{QA} mechanism for heterogeneous, multi-center \ac{DCE} breast \ac{MRI}, we designed the
experimental pipeline shown in Figure~\ref{fig:pipeline}.

\subsection{Anomaly Dataset}
\label{sec:anomaly_dataset}

We construct an anomaly dataset that enables a comprehensive and practically relevant evaluation of \ac{AD} and \ac{OOD} detection methods. 
The dataset is designed to cover a wide range of representative, structurally diverse, and realistic anomalies that reflect deviations from the normal data distribution that may occur during both model training and model inference.
The three splits consist of 854 training, 70 validation, and 496 test samples.
For details on the composition of normal, external, and anomalous samples within the test set, refer to Figure~\ref{app:data_split_test}.
Training and validation datasets for \ac{AD} and \ac{OOD} detection model development exclusively contain samples that conform to the definition of normality, ensuring that models learn only the normal data distribution without exposure to anomalous samples. 
In contrast, the test dataset contains a diverse set of anomalous samples in addition to semantically similar samples from external institutions as well as "normal" data from the same distribution as the training and validation datasets.

\paragraph{Normal Data}

Normal data is defined as unilateral \ac{DCE} breast \ac{MR} subtraction images. 
All images showing breast implants or post-mastectomy cases are excluded from this definition, as both events result in noticeable anatomical alterations. 
This ensures a consistent and homogeneous representation of normality and provides us with an additional category of anomalous data.

\paragraph{External Data}

The external dataset consists of unilateral \ac{DCE} breast \ac{MR} subtraction images acquired at external institutions that are not included in the training or validation datasets. 
These samples are considered normal but represent a domain shift due to differences in acquisition protocols, scanners, or clinical practices.
The external dataset simulates unseen institutional data encountered during deployment. 
Method behavior on this dataset is evaluated separately from \ac{AD} performance, with the goal that external normal samples are not flagged as anomalous.

\paragraph{Anomaly Data}

Anomalies are defined as samples that deviate from the normal data distribution along one or more dimensions. 
This includes breast scans acquired using different imaging modalities, \ac{DCE} \ac{MR} subtraction images of different anatomical regions outside the breast region, unilateral breast \ac{MRI} acquired with incorrect or inconsistent imaging protocols, structural or anatomical alterations such as implants or mastectomy, and artifacts or errors introduced during data pre-processing. 
We also include both spatial transformation errors and inconsistencies introduced during the subtraction image generation process.

\subsubsection{Datasets}

\noindent Six publicly available medical imaging datasets were used.

\paragraph{ODELIA Breast MRI}

The primary dataset we used is the \ac{ODELIA} dataset published in 2025, a multi-center breast MRI dataset curated by the \ac{ODELIA} consortium within the European Horizon initiative on decentralized medical \ac{AI}. 
It was collected between 2006 and 2024 across six clinical institutions in five European countries: \ac{UKA} in Germany, \ac{CAM} in the United Kingdom, \ac{MHA} in Greece, \ac{RUMC} in the Netherlands, \ac{UMCU} in the Netherlands, and \ac{RSH} in Spain. It includes three diagnostic labels per breast: no lesion, benign lesion, and malignant lesion~\citep{MuellerFranzes2025a, Odelia2025Challenge}.

\paragraph{Duke Breast Cancer MRI}

We used a subset of patients without mastectomy and implants from the \ac{DUKE} dataset as part of the external dataset. 
This single-center dataset was originally collected by Duke University School of Medicine, Durham (United States), from 2000 to 2014. 
It was intended to investigate the extent of disease in breast cancer patients using MRI and included patients with at least one breast having a malignant lesion~\citep{Saha2021}.

\paragraph{QIN-BREAST CT}

A subset of chest \ac{CT} images is used from the QIN-BREAST dataset published in 2016. 
The data was originally provided by Vanderbilt University (United States) at multiple treatment time points to investigate treatment response in breast cancer patients~\citep{Li2016}.

\paragraph{TCGA-KIRP, TCGA-LIHC, and Prostate Fused MRI Pathology}

Lastly, subsets of patients with available pre- and post-contrast agent images are included from the following three public datasets: the \ac{TCGA-KIRP}~\citep{Linehan2016} from 2016, the \ac{TCGA-LIHC}~\citep{Erickson2016} from 2016, and the \ac{PFMRIP} dataset~\citep{Madabhushi2016} from 2016, with these three datasets providing \ac{DCE} \ac{MRI} from kidney, liver, and prostate scans, respectively.

\subsubsection{Data Pre-Processing}

All images were resampled to match the input size of the original \ac{ODELIA} unilateral \ac{DCE} breast \ac{MR} subtraction images $(256, 256, 32)$.
Normal data includes all unilateral \ac{DCE} breast \ac{MR} subtraction images from the original training and validation datasets provided in the \ac{ODELIA} challenge dataset, excluding images with either implants or mastectomies, as well as data from the \ac{RSH} hospital, which is used as part of the external dataset.
We ensured that all three diagnostic labels were represented in each split and that left and right breast images from the same patient were always assigned to the same split.
The external dataset contains the first ten patients from the \ac{DUKE} dataset and the first 15 patients from the \ac{ODELIA} \ac{RSH} dataset, excluding mastectomy and implant patients. 
All images were pre-processed identically to the original \ac{ODELIA} data using a geometry- and threshold-based cropping algorithm\footnote{\url{https://github.com/mueller-franzes/odelia_breast_mri}, commit \texttt{6c79876}, accessed 2025-04-09}.

\begin{figure*}
    \centering
    \includegraphics[width=\textwidth]{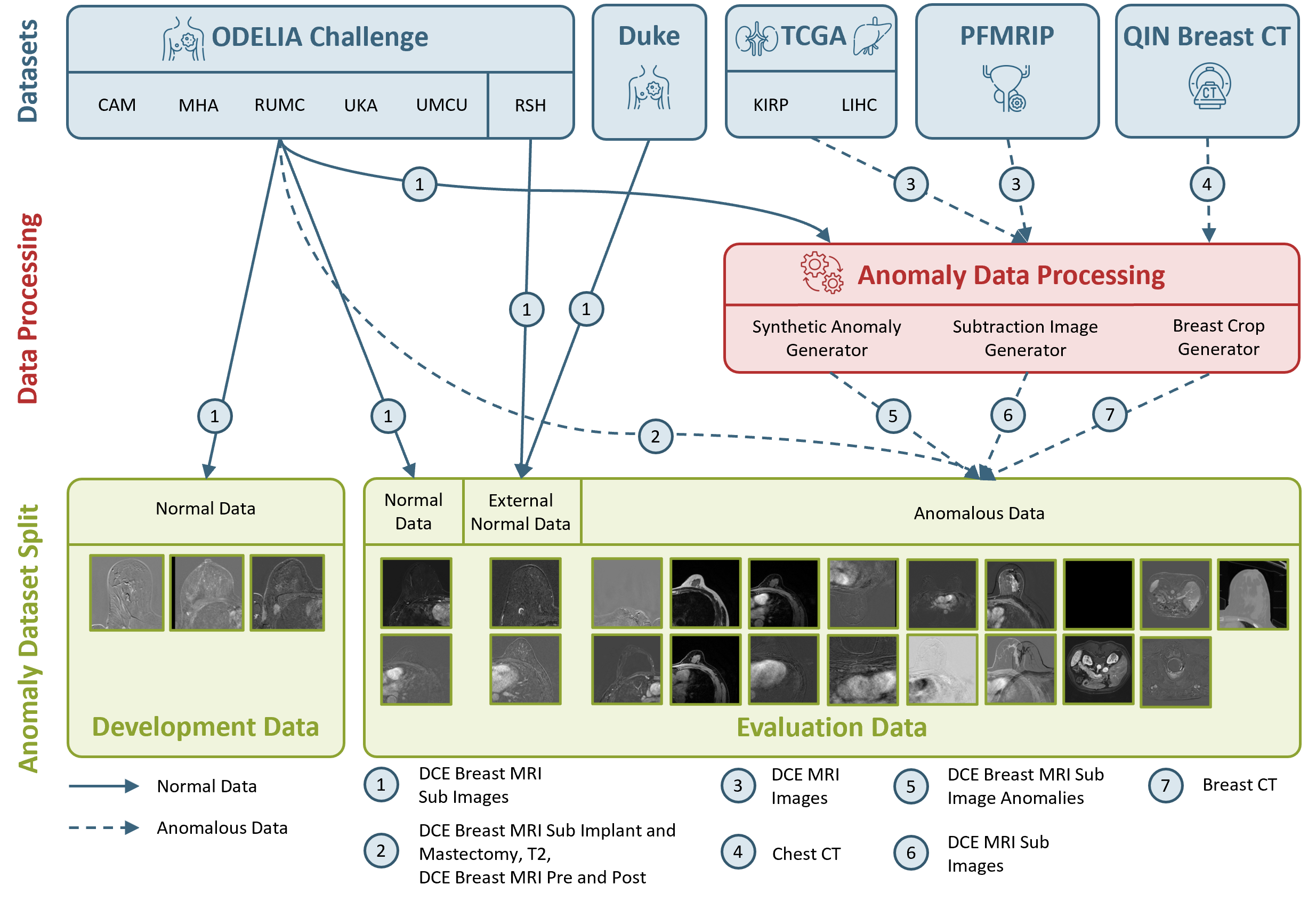}
        \caption{Data flow and anomaly construction overview for the development and evaluation data. 
        Normal \ac{DCE} breast MRI subtraction images from the \ac{ODELIA} dataset are included in the training, validation, and normal test data for both development and evaluation. 
        External normal data from \ac{DUKE} and \ac{RSH} institutions simulate unseen sites for the external normal evaluation data.
        Anomalies from the \ac{ODELIA} dataset, including implants, mastectomy, and protocol variants (T2, pre- and post-contrast \ac{DCE} \ac{MRI}), are only included in anomalous evaluation data. 
        Normal test images from the \ac{ODELIA} dataset are used to generate subtraction image inconsistencies and geometrical transformations.
        \ac{DCE} \ac{MRI} from anatomically distinct regions (kidney, liver, prostate) are further processed to generate \ac{DCE} \ac{MR} subtraction images. 
        Chest \ac{CT} images are cropped to the breast region to simulate a different imaging modality for anomalous evaluation data.
        All samples are resampled to the \ac{ODELIA} input resolution $(256, 256, 32)$. The arrows mark data flows from the source to the evaluation split.}
    \label{fig:data_examples}
\end{figure*}

Five datasets were used to simulate samples for the anomaly dataset. 
The first ten QIN-BREAST \ac{CT} images were cropped to a region comparable to the breast \ac{MRI} field of view to represent a different imaging modality, resulting in 20 unilateral images.
\ac{DCE} \ac{MR} subtraction images of the kidney, liver, and prostate were generated from the \ac{TCGA-KIRP}, the \ac{TCGA-LIHC}, and the \ac{PFMRIP} datasets to simulate incorrect anatomical regions. We included the first ten patients with available pre- and post-contrast agent images from each dataset.
From the \ac{ODELIA} dataset, the patients from the test split were used to generate the following anomalies.
Incorrect imaging protocols were simulated using T2-weighted sequences and pre- and post-contrast agent \ac{DCE} \ac{MRI}.
Spatial transformation errors were modeled as bilateral images, bilateral images cropped to the center, not the breast, breasts with wrong cropping and wrong translation, as well as vertical flipping of normal test images.
The incorrect cropping and translation could shift up to 50\% of the originally visible anatomy outside the image boundaries.
Subtraction inconsistencies were modeled as wrong patient subtraction, wrong side subtraction, same image subtraction, and wrong subtraction order.
Structural alterations were represented using mastectomy and implant images.
The data processing pipeline is shown in Figure~\ref{fig:data_examples}.
Examples for normal, external, and anomalous data are illustrated in Figure~\ref{app:data_view}.

\subsection{Taxonomy of Anomalies}

We introduce a taxonomy of radiological image data anomalies designed to support systematic evaluation of \ac{AD} and \ac{OOD} detection methods from a human visual perception perspective. 
The taxonomy provides a generalized framework for categorizing the different sources of error introduced by an anomaly as well as a perceptual distance to \ac{ID} data, independent of any specific imaging modality or clinical task. 
It is intended to be transferable across medical use cases. While the dimensions themselves remain fixed, they can be used to identify and categorize anomaly types relevant to any given application, helping to identify potential sources of error during model evaluation.
The taxonomy assigns each anomaly type a grade along four dimensions, each with three degrees of impact. The first three dimensions, \textit{protocol and modality}, \textit{anatomical and structural alterations}, and \textit{orientation and field of view}, describe the cause of the anomaly, while the fourth dimension, \textit{spatial extent}, describes its impact.

\paragraph{Protocol and Modality}

This captures the overall style of the image, with grade being (0) for the same modality and protocol with technically correct execution of pre-processing, (1) for the same modality and different protocol, optionally with wrong pre-processing, and (2) for a different modality. 
In our setting, correct pre-processing refers to subtracting pre-contrast from a post-contrast \ac{DCE} \ac{MRI}.

\paragraph{Anatomical and Structural Alterations}

This describes the image content with (0) being the same organ with expected variations, (1) being the same organ with structural alterations, and (2) being a different organ or image content.
Examples of structural alterations include local pre-processing artifacts, implants, pacemakers, etc.

\paragraph{Orientation and Field of View}

This focuses on the geometrical transformations of the image. (0) refers to no geometric misalignment, with the organ being fully visible with the right orientation, (1) describes misalignment with the breast being partly visible or in the wrong orientation, and (2) describes major misalignment with the breast not being visible at all.

\paragraph{Spatial Extent}

This describes the extent of the anomaly, excluding constant background regions. 
No anomaly is graded as (0), a small artifact that affects the image locally is graded as (1), and a large artifact that affects the image globally is graded as (2).
\\

The sum of the individual grades allows a categorization into \ac{ID} (0), near-\ac{OOD} (1-2), medium-far-\ac{OOD} (3-4), and far-\ac{OOD} ($\geq$5), inspired by~\citep{Graham2022a}.
The full grading for our anomalies is described in Table~\ref{tab:anomaly-taxonomy}.
We applied it to the anomalies for the head \ac{CT} use case presented by~\citet{Graham2022a}.
This comparison serves as an initial test of the transferability of the taxonomy to a different medical use case, anatomy, and modality. 
The resulting categories assigned to the head \ac{CT} anomalies are mostly consistent with the \ac{OOD} categories reported by~\citet{Graham2022a} and~\citet{Graham2023}, see Table \ref{app:taxonomy-ct}. This suggests that the four grading dimensions capture perceptually meaningful distinctions in distribution shift independent of imaging modality and anatomical region for volumetric radiological image data.
However, this comparison reflects only a single external use case and a qualitative agreement in category groupings rather than a quantitative validation.
A comprehensive evaluation of the taxonomy's generalizability across additional modalities and tasks is left for future work.

\begin{table*}
    \caption{Anomaly taxonomy and grading. Each anomaly type is graded on four dimensions with a grade of 0-2 per dimension. Aggregate grades result in four categories. 0 = \ac{ID}, 1-2 = near-\ac{OOD}, 3-4 = medium-far-\ac{OOD}, $\geq$5 = far-\ac{OOD} data. External data receives a grade of 0. By definition, it is \ac{ID} with respect to image content or near-\ac{OOD} with domain shift arising from institutional factors.}
    \label{tab:anomaly-taxonomy}
    \centering
    \begin{tabular*}{\textwidth}{@{\extracolsep{\fill}}l cccc c}
    \toprule
    Group & 
    \makecell{Protocol and\\Modality} &
    \makecell{Anatomical and\\Structural\\Alterations} & 
    \makecell{Orientation\\and Field\\of View} & 
    \makecell{Spatial Extent} &
    \makecell{Total\\Grade} \\
    \midrule
    \multicolumn{6}{c}{In-distribution} \\
    \midrule
    Normal Data&       0&       0&  0& 0&0 \\
    \midrule
    \multicolumn{6}{c}{External normal data} \\
    \midrule
    External Data&       0&       0&  0& 0&0 \\
    \midrule
    \multicolumn{6}{c}{Near-out-of-distribution} \\
    \midrule
    Vertical Flip& 0& 0&  1& 0&1\\
    \makecell[lt]{Wrong Crop +\\Wrong Translation}& 0& 0&  1& 0&1\\
    \makecell[lt]{Bilateral +\\Wrong Crop}& 0& 0&  1& 0&1\\
    Bilateral& 0& 0&  0& 1&1\\
    Implant& 0& 1&  0& 1&2 \\
    Mastectomy& 0& 1&  0& 1&2 \\
    \makecell[lt]{Wrong Patients\\Subtracted}& 0& 0&  0& 1&1\\
    \makecell[lt]{Wrong Side\\Subtracted}& 0& 0&  0& 2&2\\
    \midrule
    \multicolumn{6}{c}{Medium-far-out-of-distribution} \\
    \midrule
    \makecell[lt]{Wrong Subtraction\\Order}& 1& 0&  0& 2&3\\
    DCE Post& 1& 0&  0& 2&3 \\
    DCE Pre& 1& 0&  0& 2&3 \\
    T2& 1& 0&  0& 2&3 \\
    Breast CT& 2& 0&  0& 2&4\\
    \midrule
    \multicolumn{6}{c}{Far-out-of-distribution}\\
    \midrule
    Kidney MRI& 0& 2& 2& 1&5\\
    Liver MRI& 0& 2&  2& 1&5\\
    Prostate MRI& 0& 2&  2& 1&5\\
    \makecell[lt]{Same Image\\Subtracted}& 1& 2&  0& 2&5\\
    \bottomrule
    \end{tabular*}
\end{table*}

\subsection{Implementation and Experiments}

\subsubsection{Method Selection}

We selected four approaches, one from each of the two main categories introduced in Section \ref{sec:ad-categories} and two hybrid approaches, with the primary selection criterion being image-level \ac{AD} performance as demonstrated in prior work.
From the category of projection-based methods, we selected the memory bank-based method \ac{PC}, which demonstrated promising overall performance in two benchmarks for industrial and medical \ac{AD}~\citep{Bao2024a, Liu2024, Roth2022a}.
While none of the reconstruction-based methods in the medical image-focused \ac{AD} benchmark described by~\citet{Bao2024a} had comparable performance to the investigated memory bank-based methods \citep{Bao2024a}, we included an approach by \citet{Bercea2024a} who proposed a two-dimensional \ac{RA} approach specifically designed and validated for multiple medical imaging modalities~\citep{Bercea2024, Bercea2024a}.
We extended these two approaches with domain-specific adaptations, as they were not originally designed to process volumetric radiological image data.
Furthermore, we included two hybrid \ac{OOD} detection approaches for volumetric radiological image data. 
Both approaches combine discrete latent space compression using reconstruction-based models with either sequential density estimation or a \ac{DDPM} and were originally developed for brain \ac{CT}~\citep{Graham2022a}.

\subsubsection{Projection-based Method (PatchCore)}

For the category of projection-based approaches, we employed the memory bank-based method \ac{PC}\footnote{\url{https://github.com/amazon-science/patchcore-inspection}, commit \texttt{fcaa92f}, accessed 2025-04-30}, which was originally introduced by~\citet{Roth2022a} for industrial \ac{AD}. 
\ac{PC} extracts locally aware patch feature embeddings for normal samples from multiple hierarchical layers of a pretrained feature extractor and stores these representations in a compressed memory bank using coreset subsampling, where the memory bank is reduced to a percentage of the original number of patches.
During inference, the embeddings of new samples are compared against the embeddings in the memory bank. 
The anomaly score for a sample is the maximum distance to the nearest stored embedding across all patches.
This enables the detection of localized defects without requiring anomalous training data and supports the generation of spatial anomaly maps. 
These maps provide explainability by highlighting the specific regions in which the model identifies deviations from normal patterns~\citep{Roth2022a}.

\paragraph{Medical Feature Extractor}

The original \ac{PC} relies on models pretrained on natural images, since it was initially developed for industrial \ac{AD}. 
To obtain modality-specific and anatomically meaningful representations, we replaced the feature extractor with a medical foundation model\footnote{\url{https://github.com/FraunhoferMEVIS/MedicalMultitaskModeling}, commit \texttt{1ec98f1}, accessed 2025-12-23} introduced by~\citet{Schaefer2024} and~\citet{Nicke2025}, which won the MICCAI 2025 Lighthouse UNICORN Challenge~\citep{Unicorn2026Challenge}.

\paragraph{Positional Encoding Extension}

\ac{PC} does not model global spatial context.
In radiological imaging, consistent anatomical positioning across acquisitions creates strong spatial priors that can be leveraged to improve \ac{AD} performance.
In our dataset, if properly pre-processed, the thorax is consistently located near the bottom edge of the image in the axial view. At the same time, the breast is approximately centered, see Figure ~\ref{app:data_view}.
We incorporated these priors by concatenating a normalized \ac{PE} $[\alpha_{\mathrm{pos}}x,\, \alpha_{\mathrm{pos}}y,\, \alpha_{\mathrm{pos}}z]$ to each patch embedding, where $\alpha_{\mathrm{pos}}\in\mathbb{R}_{+}$ controls the relative influence of the spatial position over patch feature embeddings and $x, y, z \in [0,1]$ represent the relative patch position within the image. 
This modification affects both per-patch anomaly scoring and memory bank subsampling and may enhance sensitivity to position-dependent anomalies such as bilateral images, wrong cropping, in addition to wrong translation and vertical flipping.

\paragraph{Experimental Setup}

We established a stable baseline using the settings described in the original paper~\citep{Roth2022a}.
To confirm the settings from the original paper for the medical feature extractor, we also systematically varied the selected feature maps, patch size, and stride.
This did not result in consistent improvements over the baseline configuration.
Our final configuration used feature maps from layers two and three, a patch size of $(7,7)$, a stride of five, and a coreset size of one percent.
The input size for the feature extractor was kept at the original image size of $(256, 256, 32)$ voxels.
For the \ac{PE} extension, we empirically set the weighting parameter $\alpha_{\mathrm{pos}} = 10.0$, see the comment in the limitations paragraph below in Section \ref{par:limitations}.
We report both the baseline and the \ac{PE} variant to quantify the contribution of spatial context.

\subsubsection{Reconstruction-based Method (Reversed Autoencoder)}

For the reconstruction-based approach, we adopted the \ac{RA} framework proposed by~\citet{Bercea2024, Bercea2024a}\footnote{\url{https://github.com/ci-ber/RA}, commit \texttt{9b7a255}, accessed 2026-03-26}.
They introduce a two-dimensional approach that operates on middle slices of brain \ac{MRI}, pediatric wrist X-ray, and chest X-ray data. 
Their proposed \ac{RA} framework uses a \ac{SI-VAE} architecture combined with a reversed loss that minimizes embedding differences between input and reconstruction at each encoder level. 
They identify anomalies by comparing reconstructed pseudo-normal images to the original input.
Regions where the reconstruction deviates from the input indicate a mismatch with the learned normal data distribution and are used to compute the anomaly scores and maps~\citep{Bercea2024, Bercea2024a}.

\paragraph{Training Objective Extension}

The original \ac{RA} framework was initially developed for dense, smooth brain \ac{MRI} anatomy using \ac{MSE} and \ac{KL} regularization. 
In contrast, \ac{DCE} breast \ac{MR} subtraction images are characterized by fine structures and sharp tissue boundaries on a near-zero background, where \ac{MSE} alone results in over-smoothed reconstructions and poor model selection. 
This resulted in suboptimal reconstruction behavior for our use case.
To address this, we extended the training objective for the encoder $E_\phi$ and the decoder $D_\theta$ with parameters $\phi$ and $\theta$ for the baseline model.

Given a normal data sample $x$, the encoder models the posterior distribution $q_{\phi}(z \mid x)$ over latent variables, from which the latent representation $z$ is sampled. 
The decoder then reconstructs the input from $z$.
We augmented the original objective with an $L_{1}$ loss $\mathcal{L}_\mathrm{L1}$, a perceptual loss $\mathcal{L}_\mathrm{PL}$, and an SSIM loss $\mathcal{L}_\mathrm{SSIM}$~\citep{Johnson2016, Zhao2017}. 
Together, these three additions guide both the encoder and decoder to produce sharper reconstructions without altering the underlying architecture, see Equation~\ref{eq:loss-function-ra-ad}.
\begin{equation}
    \label{eq:loss-function-ra-ad}
    \begin{aligned}
        \mathcal{L}_{E_\phi}^{\mathrm{RA}}\left(x, z\right) &= \operatorname{ELBO}\left(x\right) - (1/\alpha)\left(\exp\left(\alpha \cdot \operatorname{ELBO}\left(D_\theta\left(z\right)\right)\right)\right) \\
        &\quad + \lambda_\mathrm{Emb} \cdot \mathcal{L}_\mathrm{Emb}\left(E_\phi\left(x\right), E_\phi\left(D_\theta\left(z\right)\right)\right)\\ 
        &\quad + \lambda_\mathrm{PL} \cdot \mathcal{L}_\mathrm{PL}\left(x, D_\theta\left(z\right)\right) + \lambda_\mathrm{L1} \cdot \mathcal{L}_\mathrm{L1}\left(x, D_\theta\left(z\right)\right)\\ 
        &\quad + \lambda_\mathrm{SSIM} \cdot \mathcal{L}_\mathrm{SSIM}\left(x, D_\theta\left(z\right)\right) \\
        \mathcal{L}_{D_\theta}^{\mathrm{RA}}\left(x, z\right) &= \operatorname{ELBO}\left(x\right) + \gamma \cdot \operatorname{ELBO}\left(D_\theta\left(z\right)\right)\\ 
        &\quad + \lambda_\mathrm{PL} \cdot \mathcal{L}_\mathrm{PL}\left(x, D_\theta\left(z\right)\right) + \lambda_\mathrm{L1} \cdot \mathcal{L}_\mathrm{L1}\left(x, D_\theta\left(z\right)\right)\\ 
        &\quad + \lambda_\mathrm{SSIM} \cdot \mathcal{L}_\mathrm{SSIM}\left(x, D_\theta\left(z\right)\right)
    \end{aligned}
\end{equation}
The hyperparameters $\alpha\geq0$ and $\gamma\geq0$ are set to $\alpha=2.0$ and $\gamma=1.0$~\citep{Daniel2021}. 
The weight for the embedding loss was set to $\lambda_{\mathrm{Emb}}=0.005$~\citep{Bercea2024a}. The additional loss weights were set empirically to $\lambda_{\mathrm{PL}}=1.0$, $\lambda_{\mathrm{L1}}=3.0$, and $\lambda_{\mathrm{SSIM}}=1.0$.

\paragraph{Architectural 3D Extension}

The original \ac{RA} operates on two-dimensional middle slices, discarding volumetric context. Anomalies located on other slices can not be detected in this setup.
We propose an extension of the original method that incorporates volumetric context by introducing a 3D \ac{RA}.
All two-dimensional operations, such as convolutions, batch normalization, and pooling, were replaced with three-dimensional operations in the residual blocks, encoder, and decoder.
To keep memory requirements manageable for the volumetric context, we reduced the number of encoder and decoder stages from six to four.
This extension allows the model to not only leverage inter-slice contextual information but also process full volumetric data samples, since it is not assumed that anomalies only occur on middle slices.

\paragraph{Experimental Setup}

The baseline 2D \ac{RA} model uses the architecture and settings from the original paper by~\citet{Bercea2024a} in combination with the extended training objective described in Equation~\ref{eq:loss-function-ra-ad}, since the proposed original setup did not result in meaningful model training and model selection.
It was trained for 300 epochs with early stopping based on validation \ac{MSE}, a batch size of 16, and a learning rate of $5 \times 10^{-5}$.
For each volume, the middle slice was extracted and resized to $(128, 128)$ voxels.
We did not apply intensity normalization to the 98th percentile from the original setup, as this would significantly reduce the relevant signal in our data.
Model selection was performed based on the lowest validation \ac{MSE}.
The 3D \ac{RA} uses the same extended training objective and hyperparameters, with batch size reduced to 8 and full volumes rescaled to $(128, 128, 16)$ voxels as in the original implementation. 
All other settings remain the same.

\subsubsection{Hybrid Methods}

We adopted two hybrid approaches for unsupervised \ac{OOD} detection. 
Both methods were originally developed for brain \ac{CT} scans by~\citet{Graham2022a, Graham2023} and can be applied to volumetric radiological image data.
Both hybrid approaches utilize a \ac{VQ-GAN} for discrete latent compression of three-dimensional volumes~\citep{Graham2022a, Graham2023}.

\paragraph{Transformer-based OOD Detection}

The first hybrid approach\footnote{\url{https://github.com/marksgraham/transformer-ood}, commit \texttt{
c90706c}, accessed 2025-08-12} combines the \ac{VQ-GAN} with a transformer-based density estimator that models the distribution on flattened sequences of these latent representations.
Specifically, the discrete three-dimensional representation produced by the \ac{VQ-GAN} is transformed into a one-dimensional sequence, which is then processed by a transformer trained to model the distribution of the conditional probabilities by maximizing the whole image log-likelihood over normal training data. 
This method allows us to reshape and upsample the resulting likelihood estimates to generate volumetric spatial likelihood maps with the same dimensionality as the input image, enabling the visualization and localization of anomalous regions.
The anomaly score is defined as the whole image likelihood~\citep{Graham2022a}. 

\paragraph{Denoising Diffusion Probabilistic Model-based OOD Detection}

The second hybrid approach\footnote{\url{https://github.com/marksgraham/ddpm-ood}, commit \texttt{
c90706c}, accessed 2026-07-23} uses the \ac{VQ-GAN} together with a \ac{DDPM}. 
Different levels of noise are applied multiple times to the compressed latent representations that are produced by the \ac{VQ-GAN}.
The \ac{DDPM} takes these representations as input and learns to iteratively denoise them again. 
At inference, noise is added to the compressed latent representation of the original input image corresponding to a range of timesteps. The \ac{DDPM} denoises each noisy representation, yielding multiple partial reconstructions.
Each reconstruction is then decoded back to the input image space using the \ac{VQ-GAN} and compared to the original input image using both \ac{MSE} and perceptual similarity as reconstruction quality metrics.

These similarity values are z-scored using statistics from a validation set and averaged across timesteps and metrics, resulting in a single anomaly score. 
A high similarity score indicates an \ac{ID} sample.
Spatial anomaly maps are produced similarly by aggregating pixel-wise, z-scored \ac{MSE} maps from a subset of reconstructions at several timesteps.
Since these models are trained only on \ac{ID} data, they should fail to denoise \ac{OOD} data. 
Compared to the first hybrid approach, this approach focuses primarily on the quality of the \ac{VQ-GAN} image reconstructions.
Furthermore, the \ac{DDPM} can be trained on higher-resolution latent representations than the transformer architecture due to more efficient memory scaling behavior. 
Since the similarity is measured at input image resolution, this pipeline can produce higher-resolution anomaly maps~\citep{Graham2023}.

\paragraph{Experimental Setup}

Both baseline configurations use the original settings from~\citet{Graham2023}, with our input size being $(256, 256, 32)$ voxels.
The \ac{VQ-GAN} for both setups was trained with four levels and with 128 channels per layer. 
The codebook size was $1024$ with a codebook embedding dimensionality of $64$.
We kept the original loss weighting of $0.001$ for the perceptual loss, $0.01$ for the adversarial loss, and $1.0$ for the remaining loss terms.
For training, we reduced the original batch size to $8$, kept the Adam optimizer with a learning rate of $3 \cdot 10^{-4}$, and included the proposed early stopping~\citep{Graham2022a, Graham2023}.

For the transformer, we adopted a $22$-layer memory-efficient transformer proposed by~\citet{Graham2023} and ~\citet{Graham2022a}.
The attention layers had a dimensionality of $256$ with $8$ attention heads.
Model training was performed for $100$ epochs with a learning rate of $1 \cdot 10^{-4}$ ~\citep{Graham2022a, Graham2023}. 
Again, we had to reduce the batch size to $16$.
In contrast to the original implementation, we selected the model based on the best validation loss, not the training loss, as we observed signs of overfitting.
For transformer-based \ac{OOD} detection, we experimented with different architecture settings and an extended training objective, which improved reconstruction quality but did not result in better overall detection results.
For the transformer-based \ac{OOD} detection, additional architectural variants and training objectives of the \ac{VQ-GAN} were explored, including modifications to the number of channels, codebook size, learning rate scheduling, and loss weighting, which improved reconstruction quality but did not result in consistent improvements over the baseline configuration and were therefore not pursued further.

The \ac{DDPM} was constructed with three levels with channels $[128, 256, 256]$ as proposed in the original paper. 
\citet{Graham2023} proposes a noise schedule with $T=1000$ steps with a scaled linear noise schedule. 
$\beta_{0}$ and $\beta_{T}$ were set to $0.0015$ and $0.0195$, respectively, following~\citet{Graham2023}.
The training was conducted over $12,000$ epochs with a reduced batch size of $16$ and a learning rate of $2.5 \cdot 10^{-5}$ with early stopping.
For the image reconstructions, a pseudo-linear multi-step scheduler was used with 100 timesteps. 
We tested 50 evenly spaced values between 0 and 1000 for this reconstruction process~\citep{Graham2023}.

\subsection{Method Evaluation}
\label{sec:evaluation}

We ran all methods five times with different random seeds and evaluated the methods on a test set that includes normal samples, normal samples from external institutions, and a diverse range of anomalies, see Section~\ref{sec:anomaly_dataset}.
For each data sample, we computed a scalar anomaly score with each method.
To evaluate the model performance, we computed the \ac{AUROC} for each anomaly group defined in Table~\ref{tab:anomaly-taxonomy} individually against the normal data.
In addition, we reported two aggregated \ac{AUROC} metrics. 
The sample-weighted average \ac{AUROC} reflects the overall performance across the test set for all anomalous samples against the normal samples as a whole, excluding external normal data. 
In contrast, the group-weighted average \ac{AUROC} provides a measure where each anomaly group contributes equally regardless of sample count.
The group-weighted metric was used as the primary summary statistic, as it prevents large anomaly groups from dominating the overall score.

For the external dataset, we reported the \ac{AUROC} separately and excluded it from both overall metrics, as the external normal data are intended to assess transferability rather than \ac{AD} performance. 
Accordingly, we aimed for an \ac{AUROC} of 0.5 in this setting, not 1.0, since a well-generalizing model should produce indistinguishable score distributions for internal and external normal samples.
To provide additional visual context for the plots, we computed two reference thresholds: a \ac{ROC}-derived threshold that maximizes Youden's J statistic, providing the optimal separation between normal and anomalous test samples, and a percentile-based threshold corresponding to the 97.5th percentile of the anomaly scores on the normal test set, estimated using linear interpolation between neighboring samples.

In practice, to classify new samples, both thresholds should be calculated on a validation dataset.
However, compared to the \ac{ROC}-based threshold, the percentile-based threshold does not require additional labeled anomalous data, making it fully unsupervised and therefore applicable in settings where only normal validation data is available.
In our case, the calculation of both thresholds on the test dataset makes both optimistic estimates of achievable performance rather than a realistic assessment of performance in a real-world deployment setting.

\section{Results}

\paragraph{Transferability to External Normal Data}

For external normal data, an \ac{AUROC} of 0.5 indicates that the method does not distinguish external normal data from the normal training distribution, as explained in Section~\ref{sec:evaluation}.
Reconstruction-based approaches achieved the best transferability to external normal data, with both the 2D and 3D \ac{RA} models obtaining \acp{AUROC} closest to the optimum (2D \ac{RA}: 0.553, 3D \ac{RA}: 0.557).
The 3D \ac{RA} demonstrated particularly consistent behavior across the two external sources (3D \ac{RA} \ac{DUKE}: 0.558, 3D \ac{RA} \ac{RSH}: 0.556), whereas the 2D \ac{RA} showed a larger gap between both external datasets (2D \ac{RA} \ac{DUKE}: 0.641, 2D \ac{RA} \ac{RSH}: 0.494).
Importantly, the results from the 2D \ac{RA} are not fully comparable to the other methods, as it was trained and evaluated solely on middle slices.
In contrast, projection-based methods \ac{PC} and \ac{PC}+\ac{PE} produced substantially higher separability between internal and external normal data (\ac{PC}: 0.738, \ac{PC}+\ac{PE}: 0.681). 
\ac{PC} without \ac{PE} exhibits the largest deviation from the optimum of all evaluated methods.
Notably, the \ac{PE} extension reduces this sensitivity to institutional shifts for both external sites (\ac{PC} \ac{DUKE}: 0.828, \ac{PC}+\ac{PE} \ac{DUKE}: 0.762, \ac{PC} \ac{RSH}: 0.678, \ac{PC}+\ac{PE} \ac{RSH}: 0.626), indicating that patch embeddings with spatial context reduces the influence of purely local, site-specific texture statistics.
Nevertheless, for both variants, the generalization ability varied noticeably between the two different external datasets.
Furthermore, both hybrid methods demonstrated comparably high separability between normal and external normal data (Transformer \ac{OOD}: 0.737, \ac{DDPM} \ac{OOD}: 0.637), with the Transformer-based \ac{OOD} performing comparably to \ac{PC} without \ac{PE}. 
Between the two hybrid methods, the \ac{AUROC} differed substantially between the different external sites (Transformer \ac{OOD} \ac{DUKE}: 0.909 vs. \ac{DDPM} \ac{OOD} \ac{DUKE}: 0.348, Transformer \ac{OOD} \ac{RSH}: 0.622 vs. \ac{DDPM} \ac{OOD} \ac{RSH}: 0.830), see Table~\ref{tab:results-comparison}.

\paragraph{Near-Out-of-Distribution Anomalies}

Next, we examined the detection performance on near-\ac{OOD} samples with more subtle data-level irregularities.
\textit{Orientation and field of view} anomalies, such as vertical flip, wrong crop + wrong translation, bilateral + wrong crop, and bilateral images, were generally well detected by projection-based and reconstruction-based methods, with \acp{AUROC} above 0.9 for most configurations.
The addition of \ac{PE} substantially improved \ac{PC} performance on these categories, most notably for vertical flip (\ac{PC}: 0.816 vs. \ac{PC}+\ac{PE}: 0.978) and bilateral + wrong crop (\ac{PC}: 0.799 vs. \ac{PC}+\ac{PE}: 0.973).
This confirms the expected benefit of spatial context for position-dependent anomalies.
In contrast, both hybrid approaches showed inconsistent \ac{AUROC} performance on \textit{orientation and field of view} anomalies. 
The Transformer \ac{OOD} method detected bilateral + wrong crop (0.966) reliably but performed moderately on bilateral images (0.843), poorly on vertical flip (0.669), and poorly on wrong crop + wrong translation (0.464). 
The \ac{DDPM} \ac{OOD} exhibited a complementary but equally inconsistent pattern. Wrong crop + wrong translation was detected with moderate performance (0.882), while vertical flip (0.395), bilateral + wrong crop (0.548), and bilateral (0.554) were not detected reliably.

Subtraction errors, including wrong patients subtracted and wrong side subtracted, were detected reliably by pro\-jection-based, reconstruction-based, and transformer-based \ac{OOD} detection methods (\acp{AUROC} $\geq$ 0.919), reflecting that these introduce globally inconsistent patterns that can be distinguished from normal subtraction images.
The \ac{DDPM} \ac{OOD}, however, achieved substantially lower \acp{AUROC}.

Implants and mastectomy cases represent a notable exception across all methods, with substantially lower \acp{AUROC} (implant: 0.622 to 0.762, mastectomy: 0.281 to 0.835).
However, a substantial improvement can be observed for \ac{PC}+\ac{PE} for mastectomy data, as the absence of breast tissue at expected spatial positions results in a larger distance to the patch embeddings in the memory bank (\ac{PC} mastectomy: 0.679 vs. \ac{PC}+\ac{PE} mastectomy: 0.835), see Table~\ref{tab:results-comparison}.

\paragraph{Medium-Far-Out-of-Distribution Anomalies}

In contrast to the variable performance observed for near-\ac{OOD} samples, detection becomes substantially more consistent across methods for medium-far-\ac{OOD} samples.
Almost all methods reliably detected these samples, including anomalies such as wrong subtraction order, \ac{DCE} pre- and post-contrast agent breast \ac{MRI}, T2, and breast \ac{CT}, with \acp{AUROC} consistently above 0.9.
One exception is the \ac{DDPM} \ac{OOD} method, which achieved high performance on protocol- and modality-level anomalies but completely failed to identify the wrong subtraction order (0.157), a processing error rather than a modality change, see Table~\ref{tab:results-comparison}.

\paragraph{Far-Out-of-Distribution Anomalies}

This consistently\\ high detection performance continues for far-\ac{OOD} samples.
Almost all methods successfully identified far-\ac{OOD} samples, such as kidney, liver, and prostate \ac{MRI}, achieving \acp{AUROC} above 0.965 across all methods and categories except for the \ac{DDPM}-based \ac{OOD}, confirming that anatomically unrelated image data is reliably flagged as anomalous.
Another exception is the \ac{PC} configuration without \ac{PE} for the category of the same image being subtracted (0.000).
Here, this sample consistently received lower anomaly scores than normal data, being a systematic inversion rather than an inability to distinguish between the two distributions.
Since \ac{PC} without \ac{PE} operates on local patches without global context, it failed to recognize the global absence of local enhancements from the breast tissue as anomalous.
Adding \ac{PE} resolved this failure (PC+\ac{PE}: 0.979).
The Transformer \ac{OOD} method exhibited the same critical failure (0.000).
Both variants of the reconstruction-based methods and the \ac{DDPM}-based \ac{OOD} detection reliably identified this anomaly (1.000), see Table~\ref{tab:results-comparison}.

\paragraph{Dimensions of Anomaly Taxonomy}

To better understand the source of these category-specific differences, we further analyzed performance along the four individual dimensions of the proposed taxonomy.
The first three dimensions, \textit{protocol and modality}, \textit{anatomical and structural alterations}, and \textit{orientation and field of view}, describe the causes of the anomaly. In contrast, the fourth dimension, \textit{spatial extent}, describes the impact.

Within the \textit{protocol and modality} category, all improved projection- and reconstruction-based methods successfully detected anomalies with the same modalities but used different protocols, with either correctly or incorrectly executed pre-processing, as well as anomalies with different modalities, achieving grades $\geq1$ on this dimension.
In all these cases, the \textit{spatial extent} grade was $\geq2$, meaning that the anomaly affected image patches globally. 
The hybrid approaches, in particular the \ac{DDPM} \ac{OOD}, were unable to detect all anomalies from the category \textit{protocol and modality} with grades $\geq1$ for this dimension. The Transformer \ac{OOD} was unable to detect the same image subtraction sample.

For the \textit{anatomical and structural alterations} dimension, all improved methods, as well as the Transformer \ac{OOD}, detected anomalies involving different organs or image content with grades $\geq2$. Only the \ac{DDPM} \ac{OOD} was unable to detect images that show different image contents.
Furthermore, none of our methods could detect \textit{anatomical and structural alterations} within the same organ with grade $=1$, specifically in combination with the \textit{spatial extent} being $=1$. 
Since our dataset did not include any anomalies having \textit{anatomical and structural alterations} for the same organ with a global effect, we cannot report the detectability of such cases.

Within the \textit{orientation and field of view} category, all improved methods again successfully detected all anomalies with grades $\geq1$, regardless of the corresponding grade from the \textit{spatial extent} dimension.
Neither hybrid approach could match this performance.

Considered independently, the \textit{spatial extent} category showed that all our improved methods detect anomalies with a global effect reliably, having grades $\geq2$. 
The Transformer \ac{OOD} performed similarly, with the exception being the samples with the same images being subtracted for \textit{spatial extent} $\geq2$. The \ac{DDPM} \ac{OOD} showed deficiencies for the subtraction images with \textit{spatial extent} grade $\geq2$, as well as for different anatomical regions with \textit{spatial extent} $= 1$.
Artifacts or corruptions affecting regional volumetric image patches could only be detected in some cases, consistent with the category-specific results described above.

\paragraph{Overall Performance}

\begin{figure*}[p]
    \centering
    \includegraphics[width=0.8\textwidth]{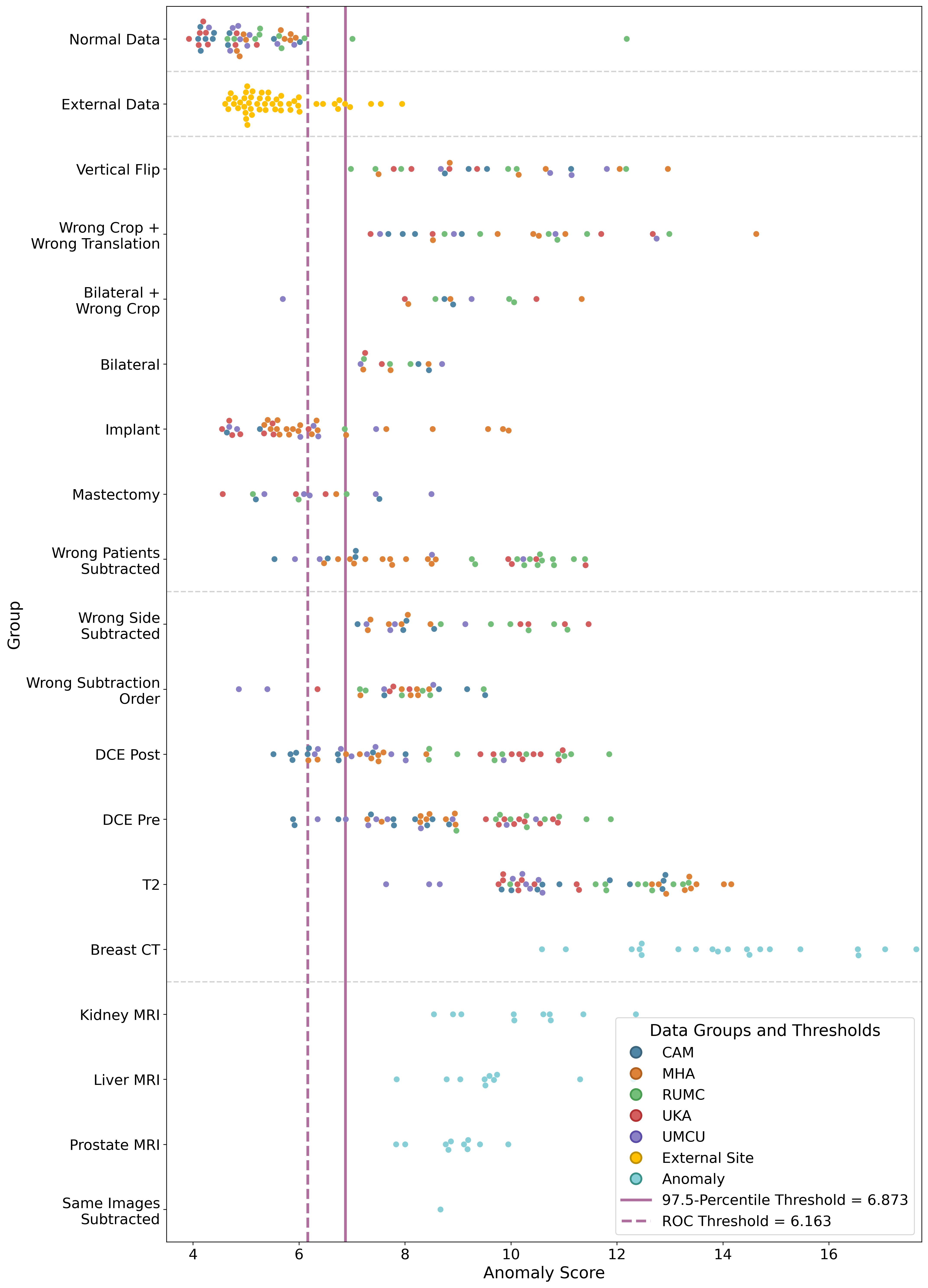}
    \caption{Anomaly score distributions for \ac{PC} with \ac{PE} on the test set. Each group corresponds to an anomaly category from Table~\ref{tab:anomaly-taxonomy}. The vertical dashed line indicates the 97.5-percentile-based decision threshold. The vertical solid line is the \ac{ROC}-based decision threshold. The horizontal dashed lines represent the separation into \ac{ID}, external, near-, medium-far-, and far-\ac{OOD} samples.}
    \label{fig:patch_core_anomaly_scores}
\end{figure*}

\begin{figure*}[t]
    \centering
    \includegraphics[width=0.8\textwidth]{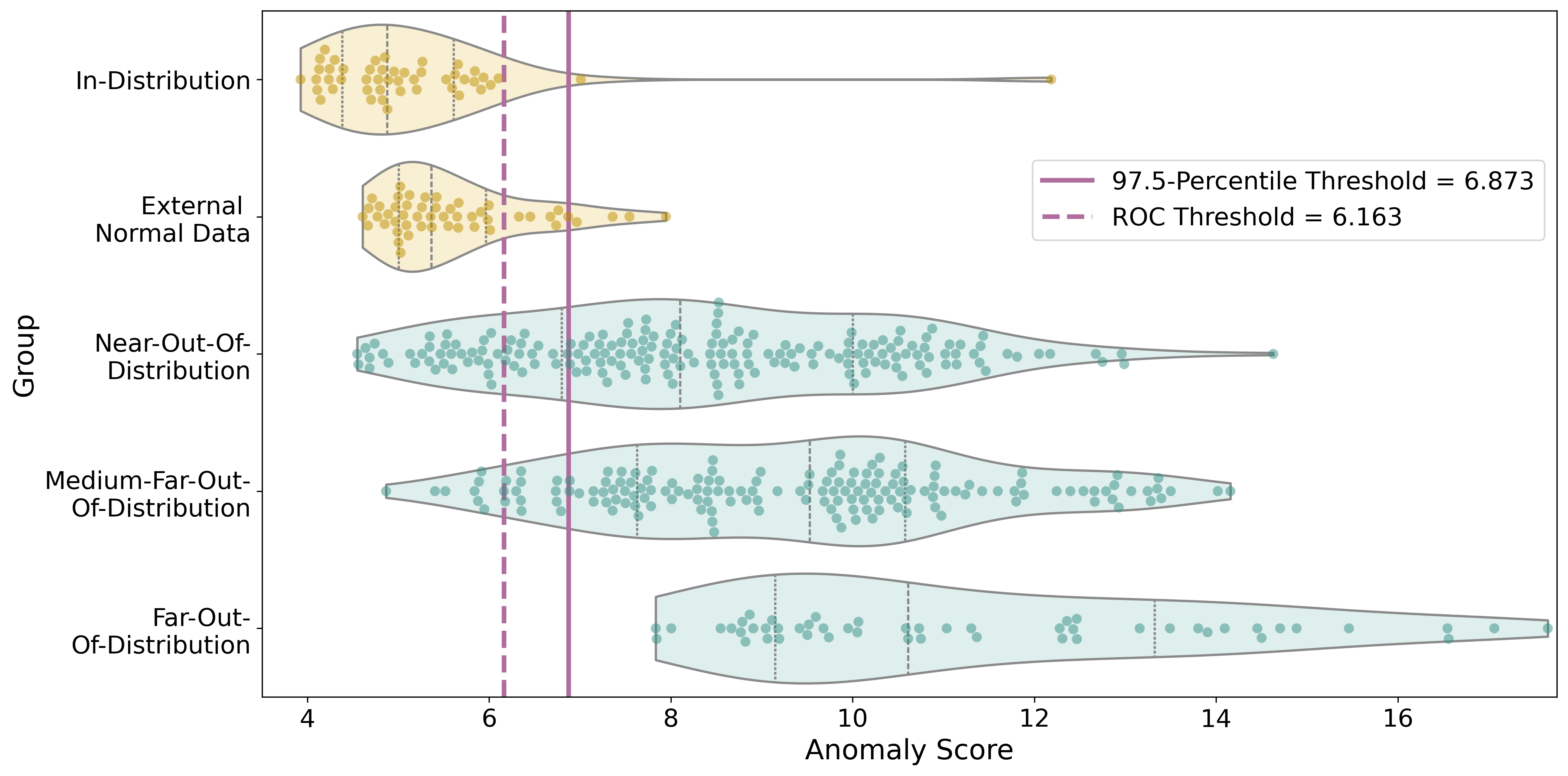}
    \caption{Anomaly score distributions per distribution category for \ac{PC} with \ac{PE} on the test set.}
    \label{fig:patch_core_score_distribution}
\end{figure*}

Overall, \ac{PC}+\ac{PE} achieved the highest group-weighted and sample-weighted performance\\ (\ac{PC}+\ac{PE}: 0.954 and 0.949), while reconstruction-based approaches demonstrated superior transferability to external normal data.
Notably, \ac{PC}+\ac{PE} improved both detection performance and external transferability relative to \ac{PC}.
The anomaly scores for all samples categorized by groups are illustrated in Figure~\ref{fig:patch_core_anomaly_scores}. 
Figure~\ref{fig:patch_core_score_distribution} shows the anomaly score distribution, grouped by \ac{OOD} level following the taxonomy.
The 3D \ac{RA}, in particular, achieved a strong combination of group-weighted and sample-weighted detection performance (3D \ac{RA}: 0.936 and 0.926) and institutional transferability (3D \ac{RA}: 0.557), supporting its suitability for cross-institutional deployment scenarios.
The Transformer \ac{OOD} method achieved moderate overall performance, which results primarily from its complete failure on same image subtraction and from poor detection of specific \textit{orientation and field of view} anomalies, as well as mastectomy cases. Furthermore, this method showed high variability between runs for almost all categories.
The \ac{DDPM} \ac{OOD} method achieved the lowest overall performance among all evaluated methods, with critical failures in all categories except for protocol and modality violations, together with the same images being subtracted.
Despite both hybrid methods being built for volumetric radiological image data, neither generalized reliably to the structural characteristics of \ac{DCE} breast \ac{MR} subtraction images without further adaptation.
Plots illustrating the anomaly scores and distribution for the 3D \ac{RA}, the Transformer \ac{OOD}, and the \ac{DDPM} \ac{OOD} are shown in the Appendix, see Figures~\ref{app:ra_3d_anomaly_scores},~\ref{app:transformer_ood_anomaly_scores},~\ref{app:ddpm_ood_anomaly_scores},~\ref{app:ra_3d_anomaly_distribution}, ~\ref{app:transformer_ood_anomaly_distribution}, and~\ref{app:ddpm_ood_anomaly_distribution}.

\begin{table*}
    \caption{Comparison of \ac{AUROC} performance on the anomaly dataset between baseline and extended methods for projection-based, reconstruction-based, and hybrid approaches. 
    Values are reported as mean $\pm$ sample standard deviation over five independent runs with different random seeds. The standard deviation quantifies run-to-run dispersion and is not a confidence interval. 
    Since \ac{AUROC} is bounded by $[0,1]$, mean $\pm$ standard deviation may formally exceed $1.0$ for means close to the upper bound, and $\pm 0.000$ indicates identical \acp{AUROC} values across all five runs.
    Each anomaly group, as defined in Table~\ref{tab:anomaly-taxonomy}, is evaluated against the normal test data.
    The 2D \ac{RA} method is only trained and evaluated on middle slices and is therefore not directly comparable to the volumetric methods. 
    For anomalous data, \acp{AUROC} $\geq$ $0.950$ are highlighted. For external normal data, \acp{AUROC} $\leq$ $0.6$ are highlighted, as an \ac{AUROC} of $0.5$ is optimal in that setting. The external normal data is excluded from both overall metrics.}
    \label{tab:results-comparison}
    \centering
    \begin{tabular*}{\textwidth}{@{\extracolsep{\fill}}l cc cc cc}
    \toprule
     & \multicolumn{2}{c }{Projection-based} & \multicolumn{2}{c }{Reconstruction-based}&\multicolumn{2}{c}{Hybrid}\\
     Groups& PC& PC+PE& 2D RA& 3D RA&Transformer OOD&DDPM OOD\\
    \midrule
    \multicolumn{7}{c}{External normal data} \\
    \midrule
    External&0.738 $\pm$ 0.011&0.681 $\pm$ 0.038& \textbf{0.553  $\pm$ 0.156}& \textbf{0.557 $\pm$ 0.013}& 0.737 $\pm$ 0.071& 0.637 $\pm$ 0.010\\
    \midrule
    \multicolumn{7}{c}{Near-out-of-distribution} \\
    \midrule
    Vertical Flip& 0.816 $\pm$ 0.020& \textbf{0.978 $\pm$ 0.001}& \textbf{0.960  $\pm$ 0.016}& \textbf{0.999 $\pm$ 0.001}& 0.669 $\pm$ 0.012& 0.395 $\pm$ 0.024\\
    \makecell[lt]{Wrong Crop +\\Wrong Translation}& 0.944 $\pm$ 0.008& \textbf{0.981 $\pm$ 0.001}& 0.936 $\pm$ 0.014& \textbf{0.971 $\pm$ 0.009}& 0.464 $\pm$ 0.082 & 0.882 $\pm$ 0.022\\
    \makecell[lt]{Bilateral +\\Wrong Crop}& 0.799  $\pm$ 0.038& \textbf{0.973 $\pm$ 0.005}& 0.922  $\pm$ 0.018& \textbf{0.958 $\pm$ 0.006}& \textbf{0.966 $\pm$ 0.040}& 0.548 $\pm$ 0.041\\
    Bilateral& \textbf{0.977 $\pm$ 0.005}& \textbf{0.970 $\pm$ 0.006}& \textbf{0.972  $\pm$ 0.014}& 0.881 $\pm$ 0.049& 0.843 $\pm$ 0.076& 0.554 $\pm$ 0.017\\
    Implant& 0.746 $\pm$ 0.009& 0.762 $\pm$ 0.008& 0.634  $\pm$ 0.012& 0.622 $\pm$ 0.028& 0.759 $\pm$ 0.073& 0.643 $\pm$ 0.008\\
    Mastectomy& 0.679 $\pm$ 0.016& 0.835 $\pm$ 0.008& 0.695  $\pm$ 0.044& 0.718 $\pm$ 0.035& 0.281 $\pm$ 0.037& 0.687 $\pm$ 0.009\\
    \makecell[lt]{Wrong Patients\\Subtracted}& \textbf{0.993 $\pm$ 0.001}& \textbf{0.966 $\pm$ 0.002}& 0.947  $\pm$ 0.006& \textbf{0.954 $\pm$ 0.012}& \textbf{0.989 $\pm$ 0.019}& 0.661 $\pm$ 0.010\\
    \makecell[lt]{Wrong Side\\Subtracted}& \textbf{0.998 $\pm$ 0.001}& \textbf{0.975 $\pm$ 0.003}& \textbf{0.964  $\pm$ 0.008}& \textbf{0.985 $\pm$ 0.010}& \textbf{0.995 $\pm$ 0.010}& 0.711 $\pm$ 0.011\\
    \midrule
    \multicolumn{7}{c}{Medium-far-out-of-distribution} \\
    \midrule
    \makecell[lt]{Wrong Subtraction\\Order}& \textbf{0.954 $\pm$ 0.006}& 0.946 $\pm$ 0.006& \textbf{0.955  $\pm$ 0.008}& \textbf{0.990 $\pm$ 0.009}& 0.919 $\pm$ 0.015& 0.157 $\pm$ 0.017\\
    DCE Post& \textbf{0.989 $\pm$ 0.001}& \textbf{0.960 $\pm$ 0.003}& 0.942  $\pm$ 0.010& 0.908 $\pm$ 0.020& 0.939 $\pm$ 0.039& 0.945 $\pm$ 0.003\\
    DCE Pre& \textbf{0.994 $\pm$ 0.001}&\textbf{0.972 $\pm$ 0.001}& \textbf{0.977  $\pm$ 0.005}& \textbf{0.977 $\pm$ 0.011}& \textbf{0.970 $\pm$ 0.035}& 0.935 $\pm$ 0.002\\
    T2& \textbf{1.000 $\pm$ 0.000}& \textbf{0.987 $\pm$ 0.001}& \textbf{0.987  $\pm$ 0.006}& \textbf{0.989 $\pm$ 0.005}& \textbf{0.967 $\pm$ 0.040} & \textbf{0.973 $\pm$ 0.004}\\
    Breast CT& \textbf{1.000 $\pm$ 0.000}& \textbf{0.996 $\pm$ 0.002}& \textbf{1.000  $\pm$ 0.000}& \textbf{1.000  $\pm$ 0.000}& 0.928 $\pm$ 0.060 & \textbf{0.981 $\pm$ 0.003}\\
    \midrule
    \multicolumn{7}{c}{Far-out-of-distribution} \\
    \midrule
    Kidney MRI& \textbf{0.997 $\pm$ 0.001}& \textbf{0.981 $\pm$ 0.001}& \textbf{0.998 $\pm$ 0.002}& \textbf{1.000 $\pm$ 0.001}& \textbf{0.989 $\pm$ 0.025}& 0.560 $\pm$ 0.033\\
    Liver MRI& \textbf{1.000 $\pm$ 0.000}& \textbf{0.981 $\pm$ 0.000}& \textbf{0.996  $\pm$ 0.005}& \textbf{0.998 $\pm$ 0.003}& \textbf{0.982 $\pm$ 0.027}& 0.482 $\pm$ 0.047\\
    Prostate MRI& \textbf{0.999 $\pm$ 0.001}& \textbf{0.977 $\pm$ 0.002}& \textbf{0.983  $\pm$ 0.017}& \textbf{0.965 $\pm$ 0.030}& \textbf{0.998 $\pm$ 0.004}& 0.231 $\pm$ 0.030\\
    \makecell[lt]{Same Image\\Subtracted}& 0.000 $\pm$ 0.000&\textbf{0.979$\pm$ 0.000}& \textbf{1.000  $\pm$ 0.000}& \textbf{1.000  $\pm$ 0.000}& 0.000 $\pm$ 0.000& \textbf{1.000 $\pm$ 0.000}\\
    \midrule
    \multicolumn{7}{c}{Overall AUROC} \\
    \midrule
    Sample-weighted& 0.936 $\pm$ 0.002& 0.949 $\pm$ 0.002& 0.925 $\pm$ 0.004& 0.927 $\pm$ 0.008& 0.867 $\pm$ 0.033& 0.725 $\pm$ 0.007\\    
    Group-weighted& 0.876 $\pm$ 0.003& \textbf{0.954 $\pm$ 0.002}& 0.934 $\pm$ 0.006& 0.936 $\pm$ 0.006& 0.804 $\pm$ 0.028& 0.667 $\pm$ 0.011\\
    \bottomrule
    \end{tabular*}
\end{table*}

\section{Discussion}

\paragraph{Method-Specific Strengths and Failure Modes}

The results confirm that no single method is superior across all evaluation criteria. 
The choice of method depends on the deployment scenario and the tolerance for specific failure modes.
\ac{PC} without \ac{PE} was the most straightforward to deploy, as it requires no architectural modification or task-specific optimization and achieves strong performance on globally anomalous and stylistically distinct samples. 
Depending on the imaging modality and task, replacing the default feature extraction model with a domain-specific feature extractor, as demonstrated here with a medical foundation model, can yield meaningful performance gains at minimal additional cost.
However, \ac{PC} without \ac{PE} failed on the same image subtraction anomaly and showed poor detection of geometric anomalies, as locally correct patches from incorrect spatial positions are already represented in the memory bank. 
For the same image subtraction sample, the method consistently detected these invalid samples as more \ac{ID} than normal data (\ac{AUROC}: 0.000). 
For a \ac{QA} pipeline, this may be more concerning than a near-chance result (\ac{AUROC}: 0.5), as this actively suppresses detection of a clinically meaningful processing error rather than failing to flag it.
Adding \ac{PE} resolved both limitations, additionally reduced sensitivity to institution-specific acquisition characteristics, and required tuning only a single additional hyperparameter $\alpha_{\mathrm{pos}}$.
This makes it a low-cost extension recommended whenever geometric consistency is a relevant \ac{QA} criterion and confirms that the failure originates from the absence of global spatial context.

Reconstruction-based methods, particularly the 3D \ac{RA}, offered the best balance between detection performance and institutional transferability.
The reconstruction mechanism implicitly encodes global image structure rather than local patch statistics, making it less sensitive to institution-specific acquisition patterns.
However, achieving this required substantially more careful adaptation than the projection-based approach. 
The original architecture, training objective, and input dimensionality, including the transition from two-dimensional middle slices to full three-dimensional representations, all required modification to handle full volumetric data and produce reconstructions of sufficient quality for reliable anomaly scoring.
This represents a practical limitation for deployment, where domain experts may not have the resources to perform this level of tuning.

Both hybrid approaches achieved substantially lower overall performance than the projection-based and recon\-struction-based methods, despite being specifically designed for volumetric radiological image data. 
In contrast to the other methods, these methods were applied without further modification. 
This finding suggests that methods developed and validated for one medical imaging modality do not necessarily transfer to different medical use cases without targeted adaptation.
Both the Transformer \ac{OOD} and the \ac{DDPM} \ac{OOD} methods were originally developed for brain \ac{CT}, which exhibits lower structural heterogeneity, a more homogeneous background, and more stable anatomical positioning than \ac{DCE} breast \ac{MR} subtraction images, which are characterized by fine anatomical structures, sharp tissue boundaries, and near-zero background signal.

The Transformer \ac{OOD} method exhibited consistent failure modes, showing high variability in the ability to detect geometric transformations as well as between runs.
Codebook quantization suppresses fine-grained spatial and geometric information, and the transformer's sequential likelihood model cannot recover the global context necessary to flag homogeneous volumes or spatially displaced content as anomalous.
Furthermore, the method failed to detect the same image subtraction case as in \ac{PC} without \ac{PE}.
The \ac{VQ-GAN} latent representation of a zero-valued subtraction image appears locally indistinguishable from the constant background regions present in normal breast \ac{MRI}, and the transformer consequently assigns high log-likelihood to the resulting latent sequence, incorrectly classifying it as \ac{ID}. 
Furthermore, we observed high variability between different runs, making it harder to train a reliable model.

The \ac{DDPM} \ac{OOD} method had even more failure modes. 
While it detected protocol- and modality-level deviations as well as the same image subtraction case reliably, it completely failed for all other cases.
This is the opposite of the expected behavior for a far-\ac{OOD} category that all other methods detected without difficulty.
As noted by~\citet{Graham2023}, reconstruction quality is a key prerequisite for effective \ac{DDPM} \ac{OOD} detection, since the core idea of this method is that the denoising model will fail on \ac{OOD} inputs. 
However, when the \ac{VQ-GAN} reconstruction quality is poor, especially compared to the 3D \ac{RA}, it produces blurry outputs that do not align with our fine-structured breast \ac{MR} subtraction images, see Figure~\ref{app:reconstruction}.

For both hybrid approaches, the multi-stage pipeline architecture with the \ac{VQ-GAN} training followed by the transformer or the \ac{DDPM}, respectively, further complicates diagnosis and correction of such failure modes, as errors introduced in the first stage propagate and interact with the second stage in non-transparent ways, making targeted adaptation substantially more resource-intensive than for single-stage approaches.
Adapting these methods to a new imaging domain would require substantial retuning of the \ac{VQ-GAN} architecture, codebook size, and loss weighting. 
Furthermore, we noticed signs of overfitting on the validation dataset during transformer training, indicating that testing different training strategies and loss functions should be considered.
The \ac{DDPM} model has to be considered as well, since it reduced the reconstruction quality even further when combined with the \ac{VQ-GAN} model.
This results in considerably higher engineering costs compared to the targeted modifications applied to the projection-based and reconstruction-based methods, further complicated by the multi-stage nature of these methods.

\paragraph{Implant and Mastectomy Anomalies}

For implant and mastectomy cases, \acp{AUROC} remained consistently low across all methods.
Despite explicitly excluding these cases from the definition of normality during training, the failure of all evaluated methods to reliably detect these samples as anomalies suggests that local patch-level and global reconstruction-based representations are insufficiently sensitive to the specific structural changes introduced by these two anomaly categories.
The \ac{PE} extension partially improved mastectomy detection, suggesting that the spatial absence of breast tissue is a detectable signal when explicit position information is provided.
Nevertheless, it should be noted that mastectomies and, in particular, implants can be more difficult to identify in subtraction images than in other protocols or sequences, such as pre- and post-contrast agent images before subtraction or T2-weighted images.
Representative examples showing both good and poor implant visibility in subtraction images are provided in Figure~\ref{app:implants}.
Furthermore, the high anatomical variability across patients, combined with the geometry- and threshold-based cropping algorithm from the \ac{ODELIA} dataset, may result in partial truncation of breast tissue and inconsistent positioning of the breast, meaning the breast occupies variable positions and proportions within the image. 
This may prevent \ac{AD} methods that implicitly rely on spatial regularity, a limitation that may be partially addressable through improved, anatomy-aware breast cropping through pre-processing.

\paragraph{Anomaly Taxonomy for Evaluation}

The method-specific strengths and failure modes described above represent precisely the problem that the proposed taxonomy has been designed to solve. 
Without a structured taxonomy for characterizing anomaly types, it is difficult to interpret the observed performance differences between methods, since a single aggregated \ac{AUROC} value combines anomalies that differ significantly.
The aggregate grade maps each anomaly type to a level of perceptual distance from \ac{ID} data, explaining why detection is consistently strong for medium-far- and far-\ac{OOD} anomalies and consistently weaker for some near-\ac{OOD} cases such as implants and mastectomy.
In addition, the four individual dimensions, \textit{protocol and modality}, \textit{anatomical and structural alterations}, \textit{orientation and field of view}, and \textit{spatial extent}, enable a more in-depth diagnosis of failure modes.
Our taxonomy provides a transferable framework for the systematic evaluation and comparison of methods and anomaly types across various radiological image data use cases, as evidenced by its consistent application to the head \ac{CT} anomaly categories in \citet{Graham2022a}.

\paragraph{Implications for Multi-Center and Federated Deployment}

Although external normal data may appear \ac{ID} from a task perspective, our results demonstrated that projection-based methods in particular are sensitive to institution-specific acquisition characteristics, producing \acp{AUROC} substantially above 0.5 for both external sites, whereas recon\-struction-based approaches generalized considerably better.
This holds most strongly for \ac{PC}.
\ac{PC}+\ac{PE} partially mitigates it.
This has practical implications for deploying \ac{AD} algorithms in federated and swarm learning settings, which is particularly relevant in the medical domain where multi-center collaborations are common but centralized datasets are often unavailable due to institutional data privacy restrictions.
Concretely, these settings require dedicated strategies that address how anomaly and \ac{OOD} detection methods can be developed in a distributed setup, whether models should be trained locally at each institution or collaboratively, and how to incorporate new partner institutions that join the collaboration at a later stage.

The 3D \ac{RA} and the hybrid approaches could, in principle, be trained similarly to a classification model within a federated or swarm learning framework, and the resulting model could subsequently be distributed to new partner institutions joining the collaboration. 
For \ac{PC}, a global memory bank could be constructed by aggregating locally extracted patches across sites, or locally generated memory banks could be merged into a unified global representation. 
The latter approach would also naturally extend to the integration of new institutions, whose local memory banks could be incrementally merged into the existing global memory bank.

In addition to method selection and training strategy, the external \ac{AUROC} itself provides a practical diagnostic tool. 
Evaluating a small subset of normal samples from an unseen institution and measuring the resulting separability from the training distribution gives a direct, interpretable estimate of how strongly that institution deviates from the learned notion of normality for the selected \ac{AD} or \ac{OOD} detection method. 
A value close to 0.5 indicates that the new site can be considered \ac{ID} with respect to the deployed model. 
In contrast, larger values signal a domain shift that requires investigation, e.g., retraining with local data, extending the memory bank, or applying site-specific normalization before relying on the model for \ac{QA} at that institution.

\paragraph{Anomaly and Out-of-Distribution Detection in the METRIC framework}

An additional conceptual perspective on dataset \ac{QA} is offered by the METRIC framework by~\citet{Schwabe2024}, which defines a structured set of data quality dimensions for trustworthy medical \ac{AI} and is motivated by regulatory requirements.
Our work provided algorithmic methods that cover several of these dimensions.
Specifically, the \textit{measurement process} dimension, which covers protocol error, subtraction generation inconsistencies, and spatial transformation errors, maps directly to the anomaly categories in our dataset.
The \textit{consistency} dimension, which concerns distributional shifts and protocol heterogeneity across institutions, corresponds to evaluating the methods on external data.
Our work did not address the \textit{informativeness}, \textit{representativeness}, and \textit{timeliness} dimensions, which concern label quality, temporal data drift, and variety and balance of the data.

\paragraph{Limitations and Future Work}
\label{par:limitations}

Several limitations of the present work should be noted.
First, the anomaly dataset did not include anomaly types that could be detected through non-imaging approaches, such as wrong file formats, incorrect dimensionality, or corrupted DICOM headers.
These are important in practice but are more appropriately addressed by metadata-level checks than by image-level \ac{AD}.
Second, we did not compare our methods against simpler image-level approaches such as histogram analysis or signal-to-noise ratio metrics~\citep{Galbusera2024, Herath2025}.
Therefore, the relative advantages over these lightweight approaches, particularly for near-\ac{OOD} categories such as implants or mastectomies, need to be addressed in future work.
Third, the \ac{ODELIA} challenge pre-processing pipeline introduces its own limitations. 
The rule-based breast cropping algorithm can produce misalignment or truncation due to the high anatomical variability across patients, in which the breast occupies variable positions and proportions of the field of view across subjects.
Furthermore, the resampling of the scans can introduce pre-processing artifacts that are more present in some institutions.
This may represent noise in the normal data distribution that may have affected model training and evaluation.
Fourth, the five-run evaluation provides limited statistical power for detecting small performance differences between methods.
Significance testing was not performed, and results should be interpreted accordingly.
Fifth, the generated anomalies are specifically designed for \ac{DCE} breast \ac{MR} subtraction images. 
The transferability to other data is therefore limited. 
Sixth, we did not perform tests regarding the performance of the different methods based on the dataset size, which is also an important factor to consider, especially in medical imaging, where usually less data is available.
Seventh, most evaluated methods operate on a single feature space without distinguishing between local and global representations.
PatchCore is an exception, as it aggregates features from multiple hierarchical encoder levels, but none of the evaluated methods explicitly combine local patch-level with global image-level representations.
Incorporating such multi-scale feature hierarchies, as proposed by \citet{Kadhim2026}, could improve detection performance, particularly for near-\ac{OOD} anomalies where deviations are subtle and spatially limited. 
This should be investigated in future work.
Eighth, all methods were trained and evaluated exclusively on \ac{DCE} \ac{MR} subtraction images. However, the \ac{ODELIA} dataset also provides the pre- and post-contrast agent images, as well as T2-weighted sequences. 
In practice, utilizing all available sequences, for example, by training sequence- and protocol-specific models, would likely improve detection coverage. 
This is particularly relevant for implants and mastectomy cases, which are not always clearly visible in subtraction images compared to T2-weighted, pre- and post-contrast agent images, see Figure~\ref{app:implants}.
This may partly explain the consistently low detection performance observed for these categories across all methods.
Ninth, the positional encoding weight was pragmatically determined from a set of $1.0$, $10.0$, and $100.0$ based on test set performance, whereas all other model tuning relied solely on visual inspection of the autoencoder outputs.
Future practical, more in-depth method adaptation should include a separate validation split containing anomalous samples for extended parameter tuning.
Tenth, the normal, external normal, and anomalous breast \ac{MRI} and \ac{CT} samples originate from breast imaging cohorts of female patients~\citep{MuellerFranzes2025a, Li2016, Odelia2025Challenge, Saha2021}.
Male breast \ac{MRI}, which differs substantially, is not represented in our normal data and would be expected to receive higher anomaly scores.
The \ac{PFMRIP} dataset~\citep{Madabhushi2016} is male by definition, and the \ac{TCGA-KIRP} dataset~\citep{Linehan2016} and \ac{TCGA-LIHC} dataset~\citep{Erickson2016} are of mixed sex. 
Sex is not a variable of interest for these samples, which serve solely as far-\ac{OOD} data.
Since male breast \ac{MRI} does occur in clinical practice, a \ac{QA} model deployed in clinical practice would require the normal data to be redefined accordingly.

\section{Conclusion}

This work establishes a comprehensive systematic foundation for evaluating \ac{AD} as a dataset \ac{QA} mechanism for the development and deployment of multi-center radiological imaging \ac{AI} systems, using breast cancer \ac{MRI} as a clinically relevant and technically challenging use case.
We employed \ac{AD} and \ac{OOD} detection as a \ac{QA} mechanism and introduced an \ac{AD}/\ac{OOD} detection benchmark containing seventeen \ac{QA} related anomaly categories as well as external and normal data across six public datasets.

Furthermore, we propose a four-dimensional grading-based taxonomy for systematic categorization of anomalies in radiological image data from a human visual perception perspective. 
The taxonomy supports the identification of method-specific failure modes. 
It is transferable to other radiological imaging use cases, as demonstrated through its application to an independent head \ac{CT} anomaly dataset.

Building on this benchmark and taxonomy, we introduce domain-specific extensions for two established unsupervised \ac{AD} and \ac{OOD} detection methods and test our benchmark on two additional hybrid \ac{OOD} methods specifically developed for volumetric radiological image data.
Our results demonstrate that the effectiveness of \ac{AD} and \ac{OOD} detection methods strongly depends on the degree and nature of the distribution shift, as well as the perceptual distance to the \ac{ID} data, and is further amplified by the level of domain-specific adaptation of these methods.

Medium-far- and far-\ac{OOD} samples are mostly detected reliably across projection-based, reconstruction-based, and transformer-based \ac{OOD} detection methods, while near-\ac{OOD} samples and external normal data reveal substantial method-specific differences.
Unlike the other methods, the \ac{DDPM}-based \ac{OOD} detection approach has substantial deficiencies in detecting most far-\ac{OOD} samples, likely due to a lack of domain-specific adaptation. 

Reconstruction-based approaches, in particular the proposed 3D \ac{RA}, provide the best combination of detection performance and transferability to unseen institutions.
Projection-based methods achieve strong overall detection performance with minimal setup at the cost of greater sensitivity to institutional domain shift.
This trade-off applies to \ac{PC} without \ac{PE}. 
Adding \ac{PE} mitigates it partially but does not close the gap to the \ac{RA} approaches.
Both hybrid approaches achieve substantially lower performance despite being developed for volumetric radiological image data. 
These findings underscore that methods designed and validated for a specific imaging modality do not generalize out of the box to substantially different anatomies and modalities. 

Unsupervised \ac{AD} and \ac{OOD} detection methods developed for the purpose of image dataset \ac{QA} for a specific imaging modality must, by definition, be tailored to each target application’s dedicated normal distribution.
Blanket generalization to arbitrary new applications or distributions is impossible and, indeed, undesirable. 
This applies not only to modality, anatomy, and pre-processing conventions, but also to the patient population.
The normal and external normal data in this work originates from a breast \ac{MRI} cohort of female patients, so that male breast \acp{MRI} are not represented. 
Their behaviour under the evaluated methods remains untested.
As such, the developed unsupervised \ac{AD} methods are designed to address image dataset \ac{QA} needs for diagnostic breast \ac{MRI} models developed based on the ODELIA dataset as an example application. 
They are not intended to be applied to other applications, populations, or scenarios. 
Deployment in a cohort with a different composition requires redefinition of the normal data for the intended target population.

The adaptation effort scales with pipeline complexity, making multi-stage architectures such as the two hybrid approaches with shared compression bottlenecks particularly challenging to tune for new domains.
The detection of implants and mastectomy cases remains an open challenge across all methods, representing a clinically significant limitation that future work must address.

Together, these findings demonstrate that unsupervised \ac{AD} and \ac{OOD} detection can be effectively utilized to address data integrity and dataset \ac{QA} requirements for medical \ac{AI} development and deployment, while revealing the method-specific limitations in detection performance, transferability, and sensitivity to domain mismatch that must be carefully considered when selecting and adapting these methods for real-world multi-center pipelines.

\section*{Ethics Statement}

This study used only publicly available, fully anonymized retrospective imaging data (\ac{ODELIA}, \ac{DUKE}, QIN-BREAST, \ac{TCGA-KIRP}, \ac{TCGA-LIHC}, and \ac{PFMRIP}). No new patient data was collected, and no additional ethical approval was required for this work. Ethical approval and informed consent for the original data collection were obtained by the respective data-contributing institutions as described in the original dataset publications by~\citet{MuellerFranzes2025a, Odelia2025Challenge},~\citet{Saha2021},~\citet{Li2016},~\citet{Linehan2016},~\citet{Erickson2016}, and~\citet{Madabhushi2016}.

\section*{Data Availability}

All six imaging datasets used in this study are publicly available from their original providers: the \ac{ODELIA} dataset~\citep{MuellerFranzes2025a, Odelia2025Challenge}, the \ac{DUKE} dataset~\citep{Saha2021}, the QIN-BREAST dataset~\citep{Li2016}, the \ac{TCGA-KIRP} dataset~\citep{Linehan2016}, the \ac{TCGA-LIHC} dataset~\citep{Erickson2016}, and the \ac{PFMRIP} dataset~\citep{Madabhushi2016}.
Due to the licensing terms of these source datasets, the fully pre-processed benchmark images cannot be redistributed. Instead, adapted code and data splits are publicly available at~\url{https://github.com/FraunhoferMEVIS/BreastMRIAnomalyQA}.

\section*{Funding}

This work received funding by the European Union’s Horizon Europe research and innovation programme (No. 101057091). The funder had no involvement in the study design, in the collection, analysis, and interpretation of data, in the writing of the report, or in the decision to submit the article for publication.

\section*{Disclaimer}

Views and opinions expressed are however those of the author(s) only and do not necessarily reflect those of the European Union or the European Health and Digital Executive Agency (HADEA). Neither the European Union nor the granting authority can be held responsible for them.

\section*{Declaration of Competing Interest}

The authors declare that they have no known competing financial interests or personal relationships that could have appeared to influence the work reported in this paper.

\section*{Declaration of Generative AI and AI-assisted technologies in the Manuscript Preparation}

During the preparation of this work, the authors used Claude (Anthropic) to assist with minor language editing of the manuscript text. After using this tool, the authors reviewed and edited the content as needed and take full responsibility for the content of the published article.
\printcredits

\clearpage

\appendix

\section{Supplementary Material}

\begin{figure}
    \centering
    \includegraphics[width=0.9\columnwidth]{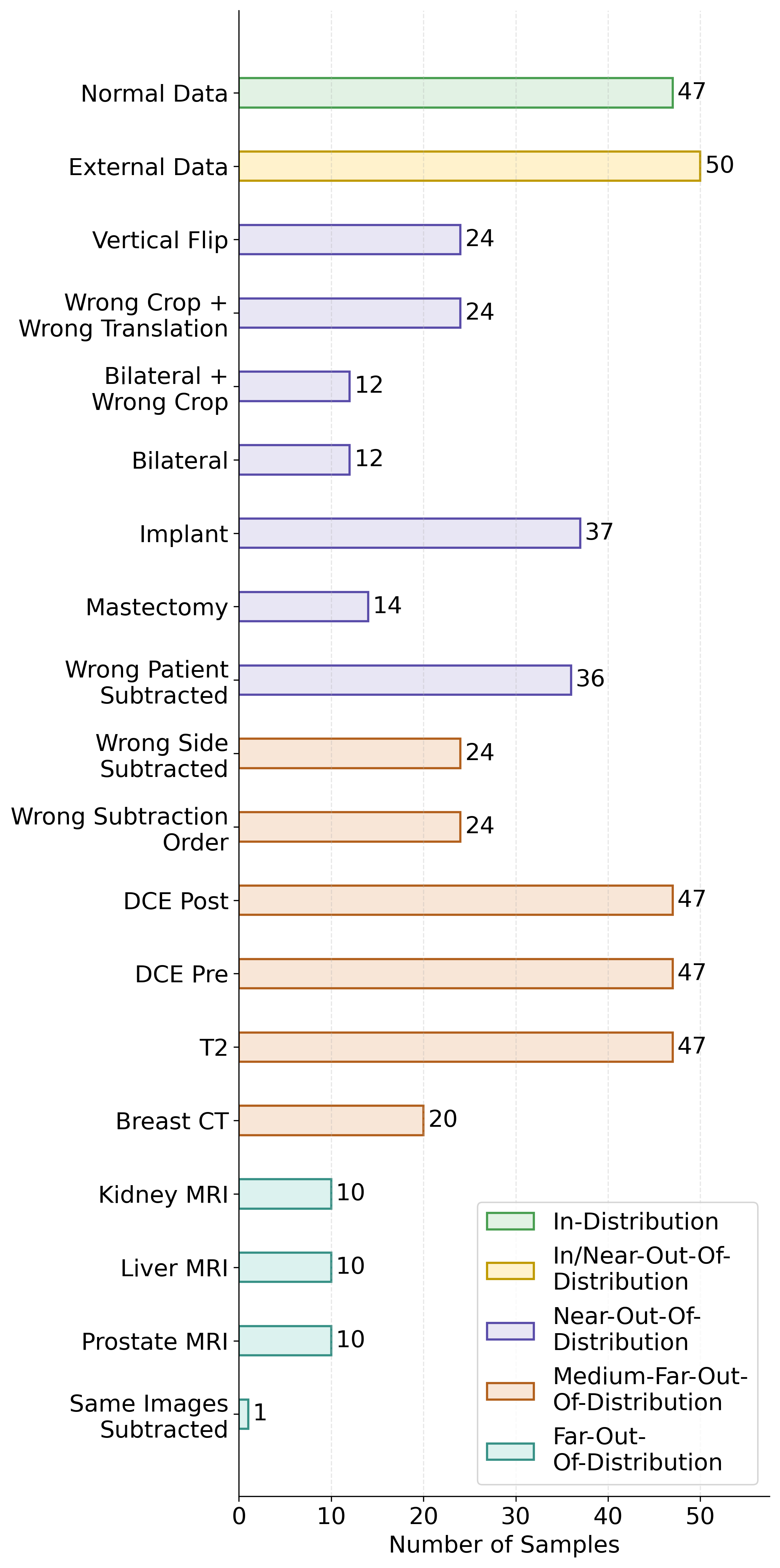}
    \caption{Detailed view of the different categories of the test set, including normal, external, and anomalous data, and their assignment to the perceptual difference to the \ac{ID} data based on the proposed anomaly taxonomy.}
    \label{app:data_split_test}
\end{figure}

\begin{figure*}
    \centering
    \includegraphics[width=0.9\textwidth]{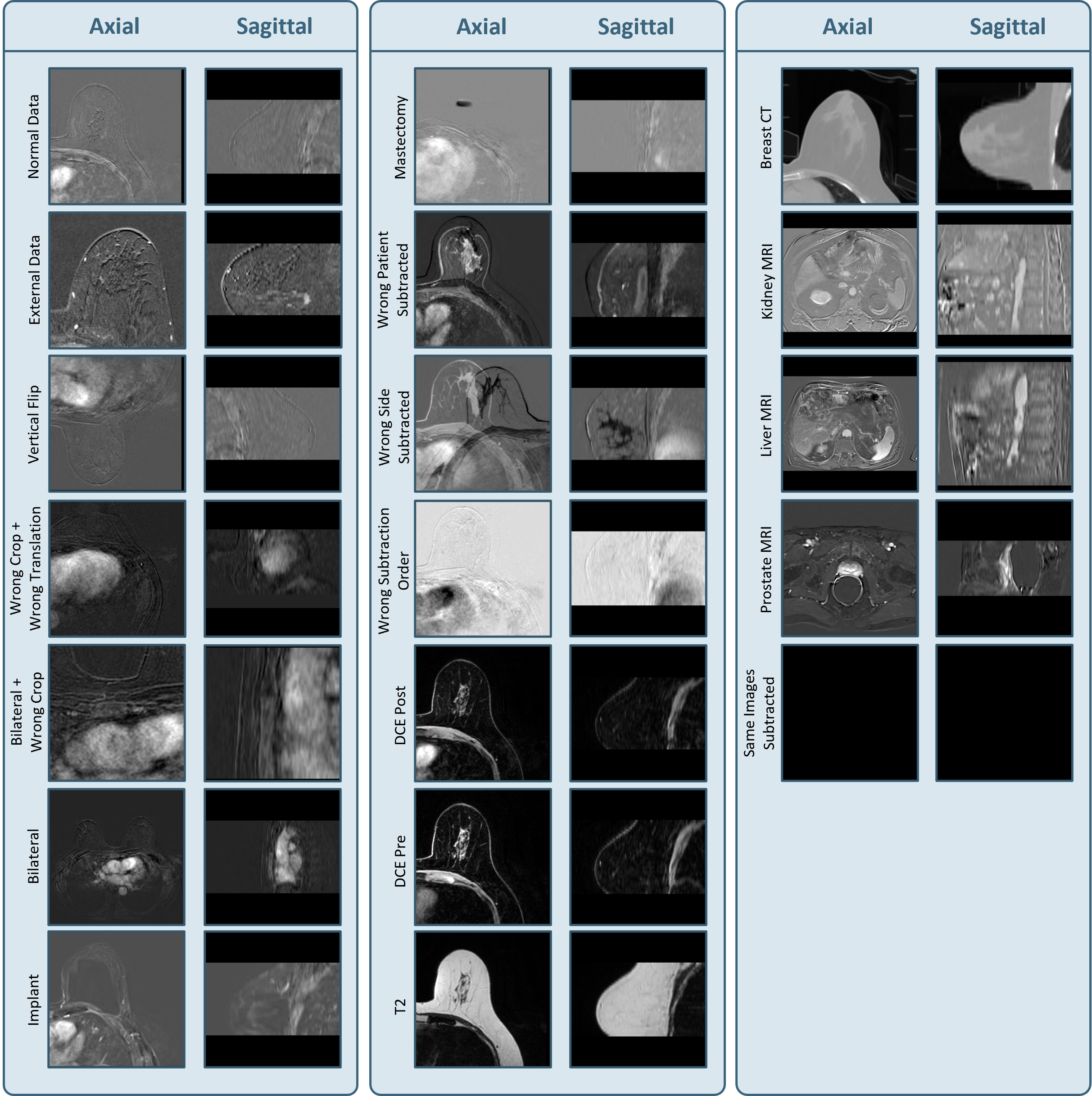}
    \caption{Examples of normal, external, and anomalous data samples. Normal data from the \ac{ODELIA} dataset include \ac{DCE} breast MR subtraction images without mastectomy or implants. Anomalies from the \ac{ODELIA} dataset include implants, mastectomy, and protocol variants (T2, pre- and post-contrast \ac{DCE} \ac{MRI}). \ac{DCE} \ac{MR} subtraction images from anatomically distinct regions (kidney, liver, prostate) are obtained from three TCGA datasets. Chest \ac{CT} images from the QIN-Breast dataset are cropped to simulate a different imaging modality. Subtraction image inconsistencies and geometrical transformations are generated from the normal test images. External normal data from \ac{DUKE} and \ac{RSH} institutions simulate unseen deployment sites.}
    \label{app:data_view}
\end{figure*}

\begin{figure*}
    \centering
    \includegraphics[width=0.8\textwidth]{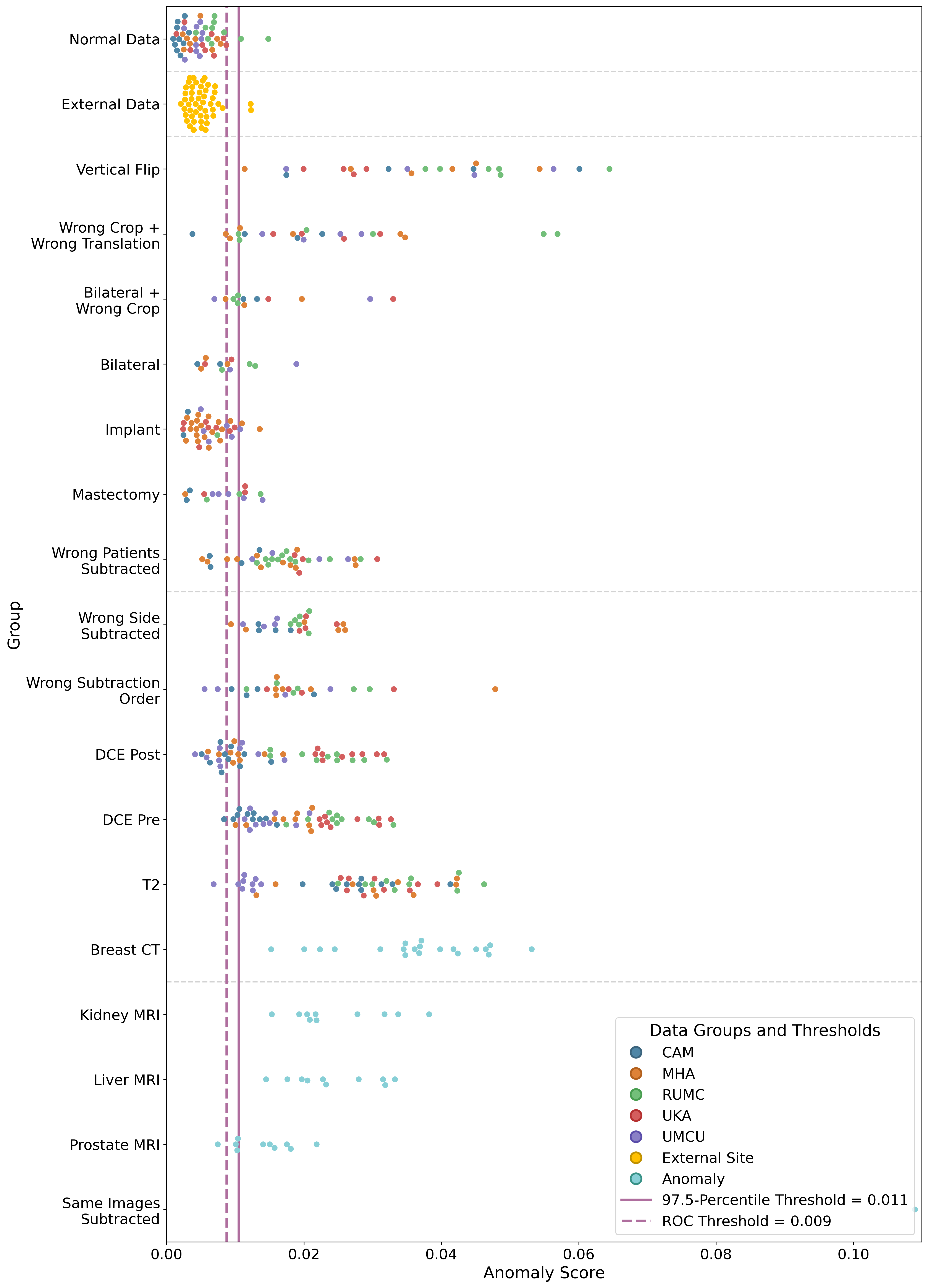}
    \caption{Anomaly score distributions for the 3D \ac{RA} on the test set. Each group corresponds to an anomaly category from Table~\ref{tab:anomaly-taxonomy}. The vertical dashed line indicates the 97.5-percentile-based decision threshold. The vertical solid line is the \ac{ROC}-based decision threshold. The horizontal dashed lines represent the separation into \ac{ID}, external, near-, medium-far-, and far-\ac{OOD} samples.}
    \label{app:ra_3d_anomaly_scores}
\end{figure*}

\begin{figure*}
    \centering
    \includegraphics[width=0.8\textwidth]{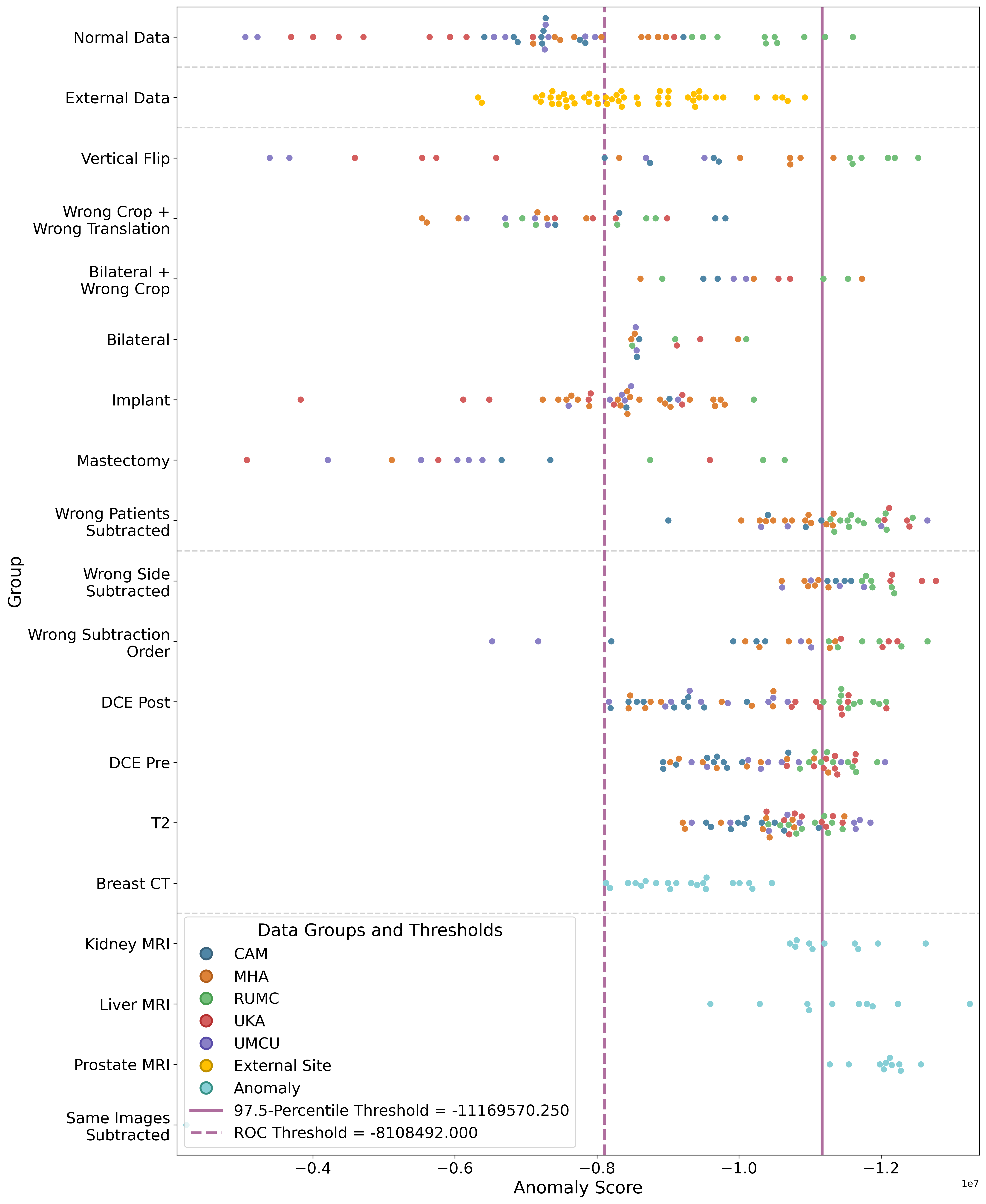}
    \caption{Anomaly score distributions for the transformer-based \ac{OOD} detection on the test set. Each group corresponds to an anomaly category from Table~\ref{tab:anomaly-taxonomy}. The vertical dashed line indicates the 97.5-percentile-based decision threshold. The vertical solid line is the \ac{ROC}-based decision threshold. The horizontal dashed lines represent the separation into \ac{ID}, external, near-, medium-far-, and far-\ac{OOD} samples.}
    \label{app:transformer_ood_anomaly_scores}
\end{figure*}

\begin{figure*}
    \centering
    \includegraphics[width=0.8\textwidth]{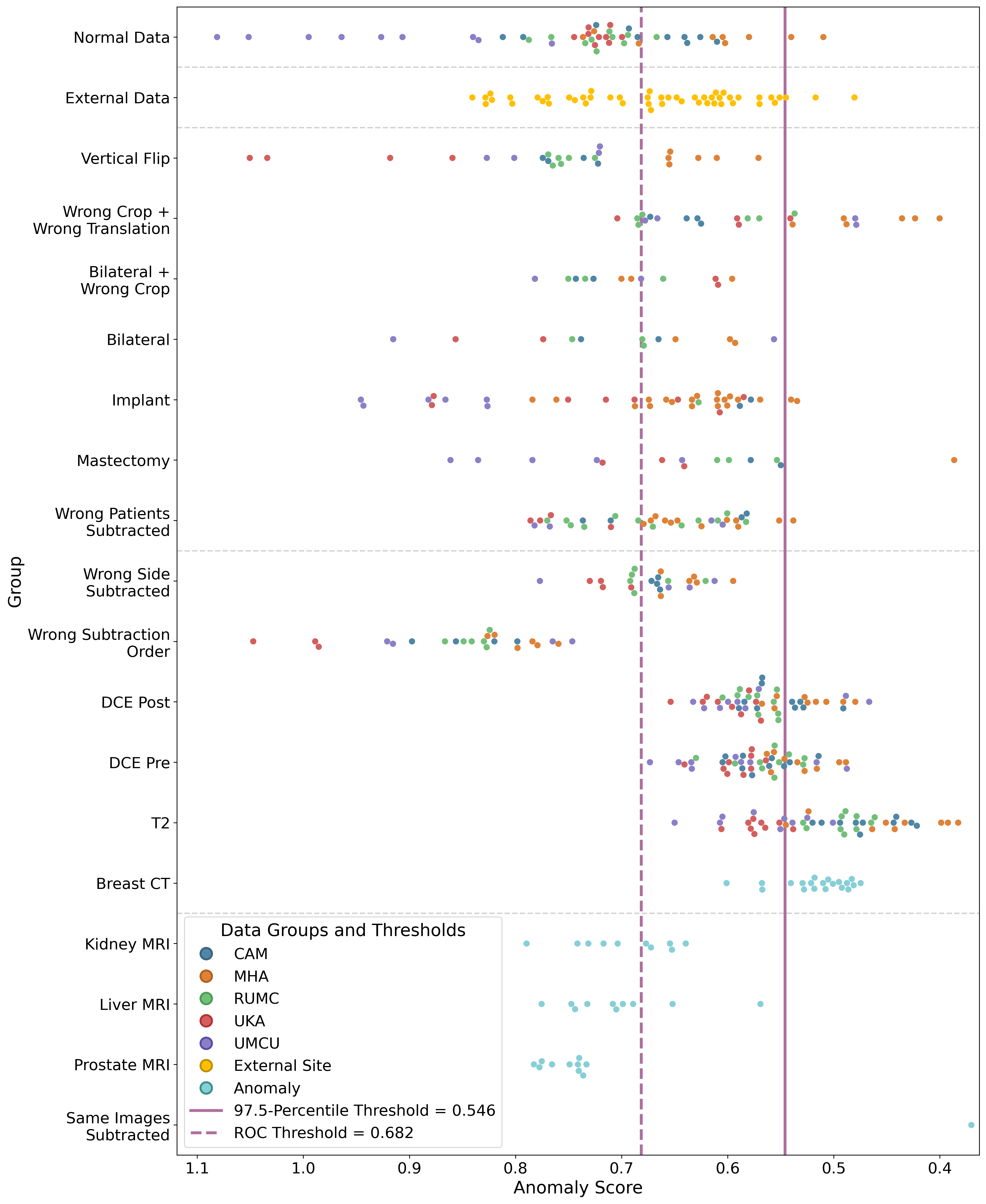}
    \caption{Anomaly score distributions for the \ac{DDPM}-based \ac{OOD} detection on the test set. Each group corresponds to an anomaly category from Table~\ref{tab:anomaly-taxonomy}. The vertical dashed line indicates the 97.5-percentile-based decision threshold. The vertical solid line is the \ac{ROC}-based decision threshold. The horizontal dashed lines represent the separation into \ac{ID}, external, near-, medium-far-, and far-\ac{OOD} samples.}
    \label{app:ddpm_ood_anomaly_scores}
\end{figure*}

\begin{figure*}
    \centering
    \includegraphics[width=0.8\textwidth]{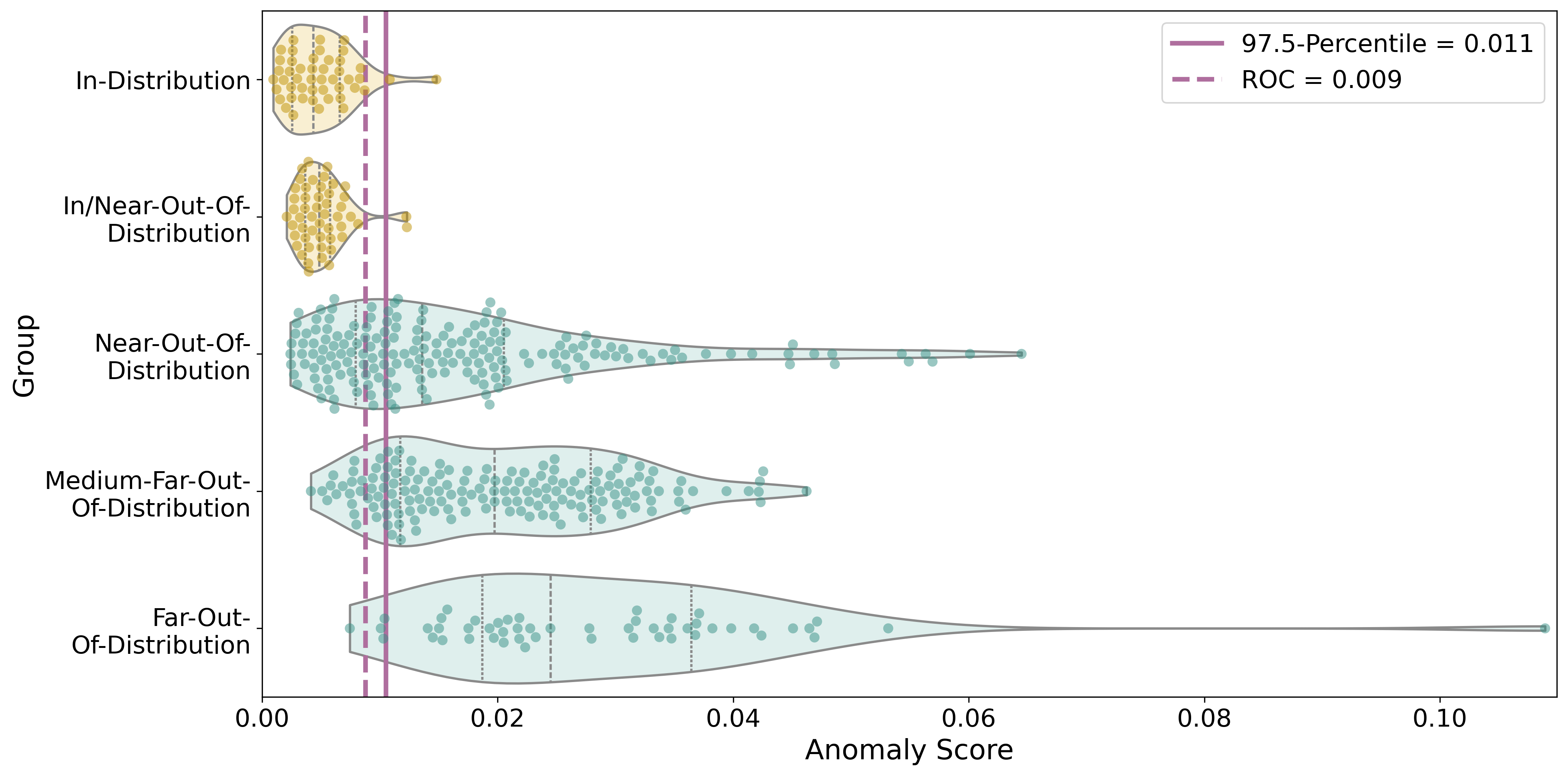}
    \caption{Anomaly score distributions per distribution category for the 3D \ac{RA} on the test set.}
    \label{app:ra_3d_anomaly_distribution}
\end{figure*}

\begin{figure*}
    \centering
    \includegraphics[width=0.8\textwidth]{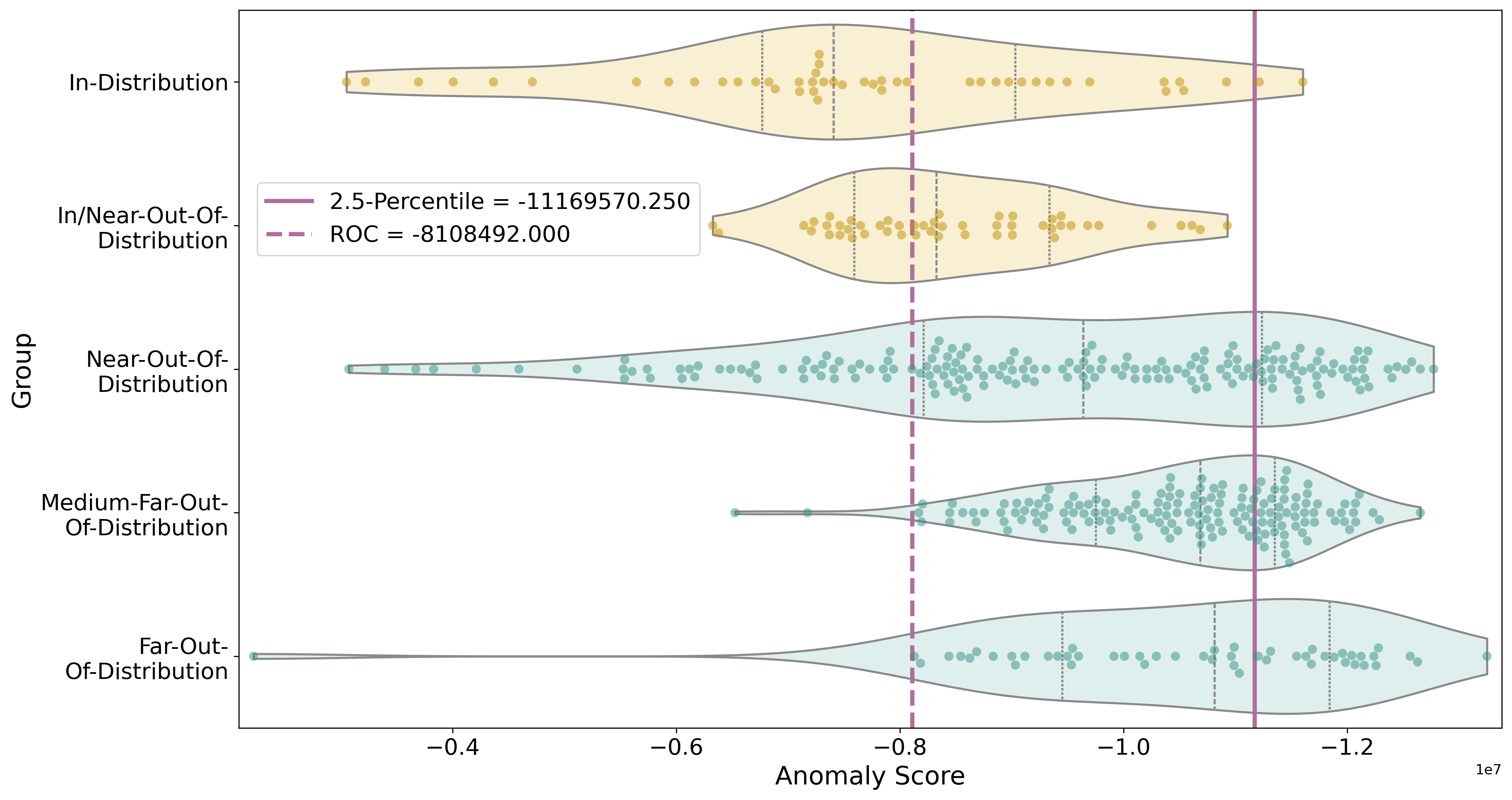}
    \caption{Anomaly score distributions per distribution category for the transformer-based \ac{OOD} detection on the test set.}
    \label{app:transformer_ood_anomaly_distribution}
\end{figure*}

\begin{figure*}
    \centering
    \includegraphics[width=0.8\textwidth]{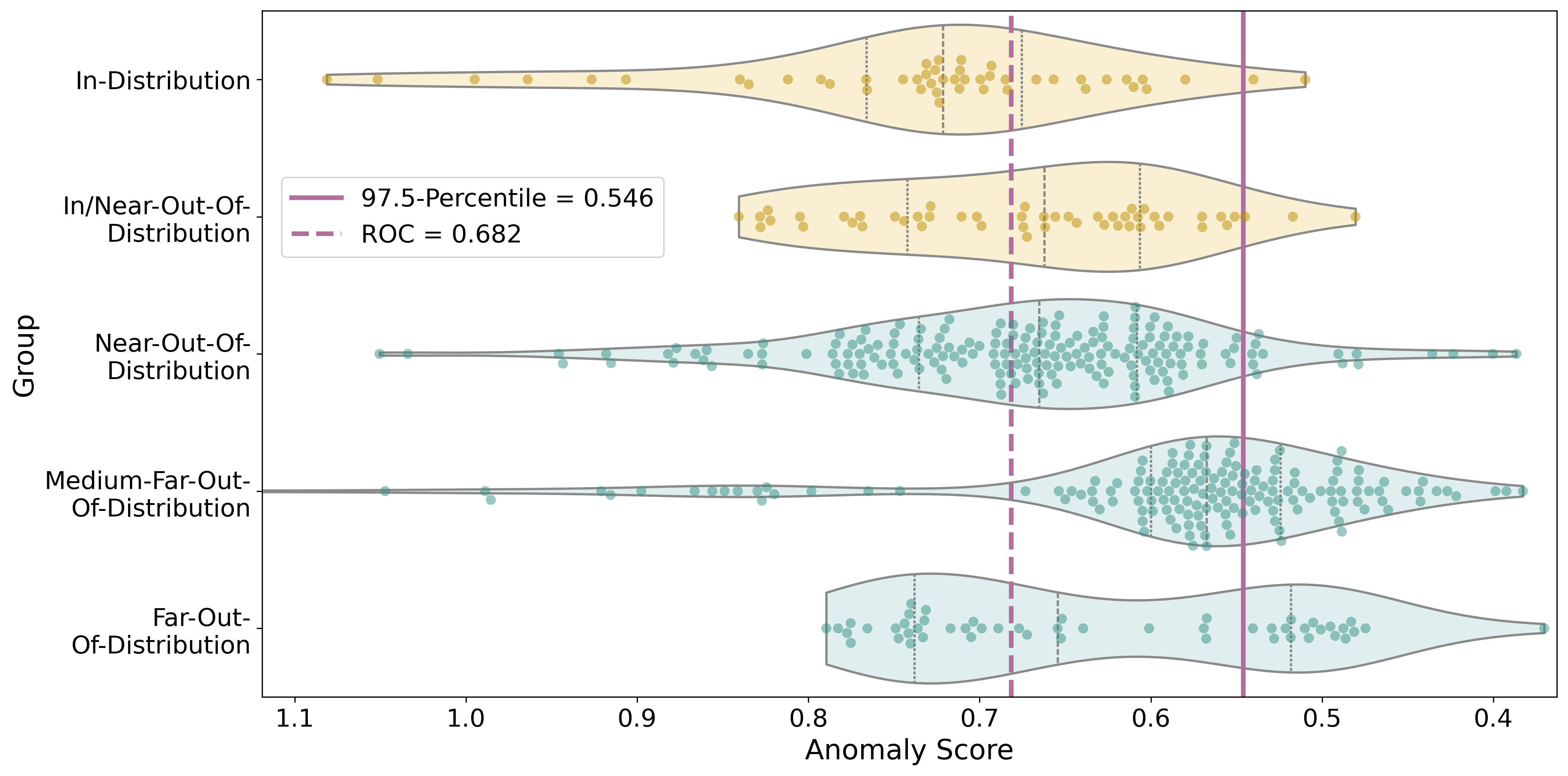}
    \caption{Anomaly score distributions per distribution category for the \ac{DDPM}-based \ac{OOD} detection on the test set.}
    \label{app:ddpm_ood_anomaly_distribution}
\end{figure*}

\begin{figure}
    \centering
    \includegraphics[width=\columnwidth]{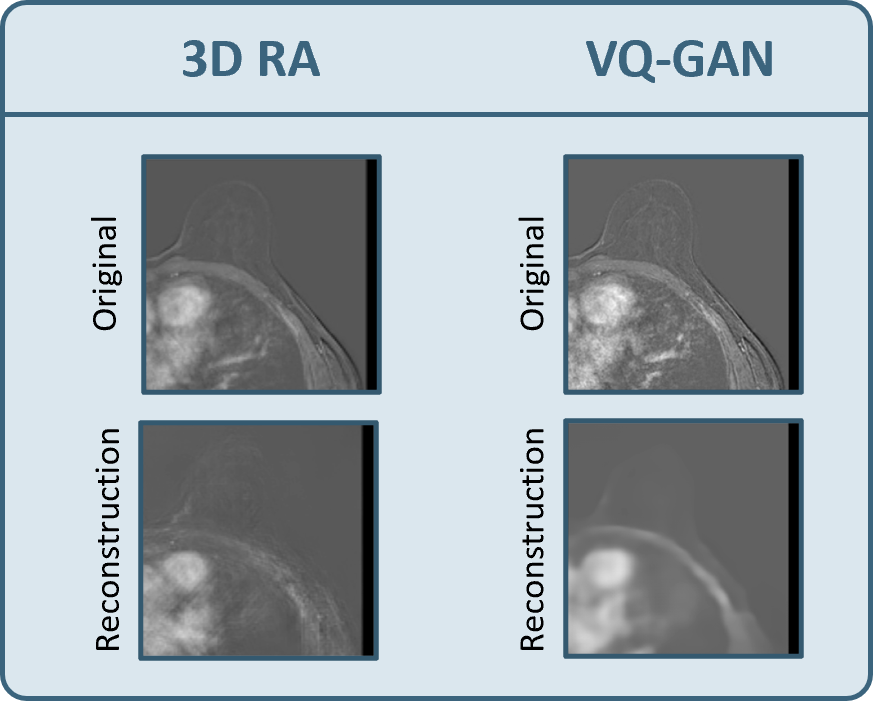}
    \caption{Qualitative reconstruction comparison for the 3D \ac{RA} and the \ac{VQ-GAN} on a representative validation sample. For each model, the first row shows the input middle slice, and the second row shows the corresponding reconstruction. The 3D \ac{RA} operates at a resolution of $(128, 128)$ voxels, while the \ac{VQ-GAN} operates on $(256, 256)$ voxels, which directly influences the small details and sharpness of the input image.}
    \label{app:reconstruction}
\end{figure}

\begin{figure}
    \centering
    \includegraphics[width=\columnwidth]{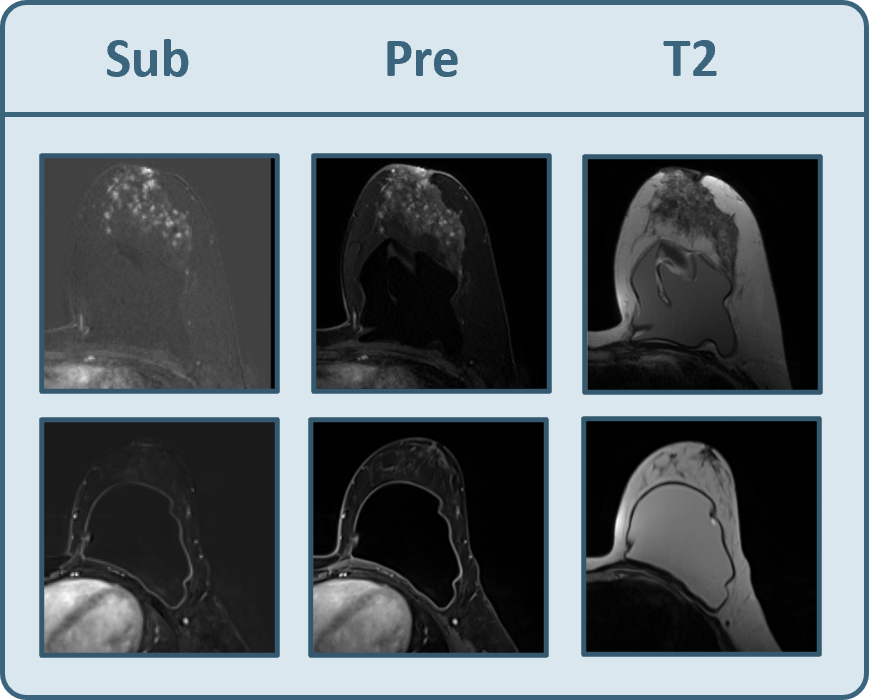}
    \caption{Two examples illustrating variability in implant visibility across \ac{MRI} sequences. Each row shows a different patient case. Columns show the \ac{DCE} subtraction image (Sub), \ac{DCE} pre-contrast agent image (Pre), and T2-weighted image (T2). Implants are more visible on T2-weighted and \ac{DCE} pre-contrast agent images than on subtraction images, where the absence of enhancement makes implant detection particularly challenging for \ac{AD} methods trained on subtraction images.}
    \label{app:implants}
\end{figure}

A detailed overview of the different subgroups within the test dataset is shown in Figure~\ref{app:data_split_test}. 
Figure~\ref{app:data_view} presents examples of normal, external, and anomalous data.
The anomaly taxonomy and grading assigned to the anomaly types defined by~\citet{Graham2023} and~\citet{Graham2022a} is shown in Table~\ref{app:taxonomy-ct}.
Figures~\ref{app:ra_3d_anomaly_scores},~\ref{app:transformer_ood_anomaly_scores},~\ref{app:ddpm_ood_anomaly_scores}, ~\ref{app:ra_3d_anomaly_distribution}, ~\ref{app:transformer_ood_anomaly_distribution}, and~\ref{app:ddpm_ood_anomaly_distribution} present the anomaly score distributions for the 3D \ac{RA}, the Transformer \ac{OOD}, and the \ac{DDPM} \ac{OOD}, respectively, per anomaly category and per distribution category for on the test set. 
Examples of reconstructions for the 3D \ac{RA} and the \ac{VQ-GAN} on a representative validation sample are provided in Figure~\ref{app:reconstruction}. 
Figure \ref{app:implants} shows the variability in implant visibility across \ac{MRI} sequences.

\begin{table*}
    \caption{Anomaly taxonomy and grading assigned to the anomaly types defined by~\citet{Graham2023} and~\citet{Graham2022a}, assuming that the brain \ac{CT} scans were acquired under a dedicated head \ac{CT} protocol, and that a different protocol was used for the \ac{CT} scans of other anatomical regions.}
    \centering
    \begin{tabular*}{\textwidth}{@{\extracolsep{\fill}}l cccc c}
    \toprule
    Group & 
    \makecell{Protocol and\\Modality} &
    \makecell{Anatomical and\\Structural\\Alterations} & 
    \makecell{Orientation\\and Field\\of View} & 
    \makecell{Spatial Extent} &
    \makecell{Total\\Grade} \\
    \midrule
    \multicolumn{6}{c}{In-distribution} \\
    \midrule
    Brain CT&       0&       0&  0& 0&0 \\
    \midrule
    \multicolumn{6}{c}{Near-out-of-distribution} \\
    \midrule
    Background (0.3)& 0& 1&  0& 0&1\\
    Background (0.6)& 0& 1&  0& 0&1\\
    Background (1.0)& 0& 1&  0& 0&1\\
    Flip L-R& 0& 0&  1& 0&1\\
    Flip A-P& 0& 0&  1& 0&1\\
    Flip I-S& 0& 0&  1& 0&1\\
    Scaling (1\%)& 0& 0&  1& 0&1\\
    Scaling (10\%)& 0& 0&  1& 0&1\\
    Skull Stripped& 0& 1&  0& 1&2\\
    Chunk top& 0& 1&  0& 1&2\\
    Chunk middle& 0& 1&  0& 1&2\\
    Noise ($\sigma$ = 0.01)& 0& 0&  0& 2&2\\
    Noise ($\sigma$ = 0.1)& 0& 0&  0& 2&2\\
    Noise ($\sigma$ = 0.2)& 0& 0&  0& 2&2\\
    \midrule
    \multicolumn{6}{c}{Medium-far-out-of-distribution} \\
    \midrule
    Head MR& 2& 0&  0& 2&4\\
    \midrule
    \multicolumn{6}{c}{Far-out-of-distribution}\\
    \midrule
    Colon CT& 1& 2& 2& 1&6\\
    Hepatic CT& 1& 2& 2& 1&6\\
    Liver CT& 1& 2& 2& 1&6\\
    Lung CT& 1& 2& 2& 1&6\\
    Pancreas CT& 1& 2& 2& 1&6\\
    Spleen CT& 1& 2& 2& 1&6\\
    Prostate MR& 2& 2& 2& 2&8\\
    Cardiac MR& 2& 2& 2& 2&8\\
    \bottomrule
    \end{tabular*}
    \label{app:taxonomy-ct}
\end{table*}



\clearpage

\bibliographystyle{cas-model2-names}

\bibliography{refs}

\end{document}